\documentclass{article} 
\usepackage{iclr2025_conference,times}

\usepackage{amsmath,amsfonts,bm}

\def\eqref#1{equation~\ref{#1}}

\def\1{\bm{1}}

\DeclareMathAlphabet{\mathsfit}{\encodingdefault}{\sfdefault}{m}{sl}
\SetMathAlphabet{\mathsfit}{bold}{\encodingdefault}{\sfdefault}{bx}{n}

\usepackage{titletoc}
\usepackage{hyperref}
\usepackage{url}

\usepackage{graphicx}
\usepackage{booktabs}

\usepackage[accsupp]{axessibility}  

\usepackage{orcidlink}
\usepackage{multirow}
\usepackage{booktabs}

\usepackage{color}
\usepackage{colortbl}

\usepackage{algorithm}
\usepackage{algpseudocode}

\usepackage{amsfonts}
\usepackage{bbm}
\usepackage{bm}
\usepackage{wrapfig,lipsum,booktabs}
\usepackage[utf8]{inputenc}
\usepackage{url}
\usepackage{booktabs}
\usepackage{amssymb}
\usepackage{bbding}
\usepackage{pifont}
\usepackage{wasysym}
\usepackage{utfsym}
\usepackage{fontawesome}
\usepackage{titlesec}
\usepackage{subfigure}

\usepackage{comment}

\titlespacing\section{0pt}{4.0pt plus 4.0pt minus 2.0pt}{0pt plus 2.0pt minus 2.0pt}
\titlespacing\subsection{0pt}{4.0pt plus 4.0pt minus 2.0pt}{0pt plus 2.0pt minus 2.0pt}
\titlespacing\subsubsection{0pt}{4.0pt plus 4.0pt minus 2.0pt}{0pt plus 2.0pt minus 2.0pt}

\definecolor{my_blue}{HTML}{d3eaf2}
\definecolor{Green}{rgb}{0.85882353, 0.90980392, 0.84705882}
\definecolor{rose}{rgb}{0.60392157, 0.53333333, 0.43921569}
\definecolor{dred}{rgb}{0.7254902, 0.09803922, 0.10588235}
\definecolor{VAIL_Green}{rgb}{0, .7, .0}

\makeatletter
\DeclareRobustCommand\onedot{\futurelet\@let@token\@onedot}
\def\@onedot{\ifx\@let@token.\else.\null\fi\xspace}

\makeatother

\hypersetup{colorlinks,breaklinks,anchorcolor=darkblue,citecolor=VAIL_Green}

\title{DebGCD: Debiased Learning with Distribution Guidance for Generalized Category Discovery}

\author{Yuanpei Liu \qquad\qquad Kai Han\textsuperscript{\textdagger}\\
Visual AI Lab, The University of Hong Kong\\
{\tt\small ypliu0@connect.hku.hk \qquad  kaihanx@hku.hk}}

\iclrfinalcopy 
\begin{document}

\maketitle
\renewcommand{\thefootnote}{\fnsymbol{footnote}}
\footnotetext[2]{Corresponding author.}

\begin{abstract}
In this paper, we tackle the problem of Generalized Category Discovery (GCD). 
Given a dataset containing both labelled and unlabelled images, the objective is to categorize all images in the unlabelled subset, irrespective of whether they are from known or unknown classes. 
In GCD, an inherent label bias exists between known and unknown classes due to the lack of ground-truth labels for the latter. 
State-of-the-art methods in GCD leverage parametric classifiers trained through self-distillation with soft labels, leaving the bias issue unattended. 
Besides, they treat all unlabelled samples uniformly, neglecting variations in certainty levels and resulting in suboptimal learning. 
Moreover, the explicit identification of semantic distribution shifts between known and unknown classes, a vital aspect for effective GCD, has been neglected. 
To address these challenges, we introduce DebGCD, a \underline{Deb}iased learning with distribution guidance framework for \underline{GCD}. 
Initially, DebGCD co-trains an auxiliary debiased classifier in the same feature space as the GCD classifier, progressively enhancing the GCD features. Moreover, we introduce a semantic distribution detector in a separate feature space to implicitly boost the learning efficacy of GCD. Additionally, we employ a curriculum learning strategy based on semantic distribution certainty to steer the debiased learning at an optimized pace. 
Thorough evaluations on GCD benchmarks demonstrate the consistent state-of-the-art performance of our framework, highlighting its superiority. 
Project page: \url{https://visual-ai.github.io/debgcd/}
\end{abstract}

\section{Introduction}
\label{sec:intro}

Over the years, the field of computer vision has witnessed remarkable progress in diverse tasks such as object detection~\cite{girshick2015fast,ren2015faster}, classification~\cite{simonyan2014very,he2016deep}, and segmentation~\cite{he2017mask,wang2020solo}. These advancements have predominantly stemmed from the availability of expansive labelled datasets~\cite{deng2009imagenet,lin2014microsoft}. However, the prevalent insufficiency of training data in real-world scenarios is a noteworthy concern. This has engendered a surge in research on semi-supervised learning~\cite{chapelle2009semi} and self-supervised learning~\cite{jing2020self}, yielding promising outcomes in comparison to supervised learning approaches. 
Recently, the task of category discovery, which was initially studied as novel category discovery (NCD)~\cite{han2019learning} and subsequently extended to its relaxed variant, generalized category discovery (GCD)~\cite{vaze2022generalized}, has emerged as a research task attracting increasing attention. 
GCD considers a partially-labelled dataset, where the unlabelled subset may contain instances from both labelled and unseen classes. The objective is to learn to transfer knowledge from labelled data to categorize unlabelled data.

\begin{figure}[ht]
\centering
\includegraphics[width = 1.0\textwidth]{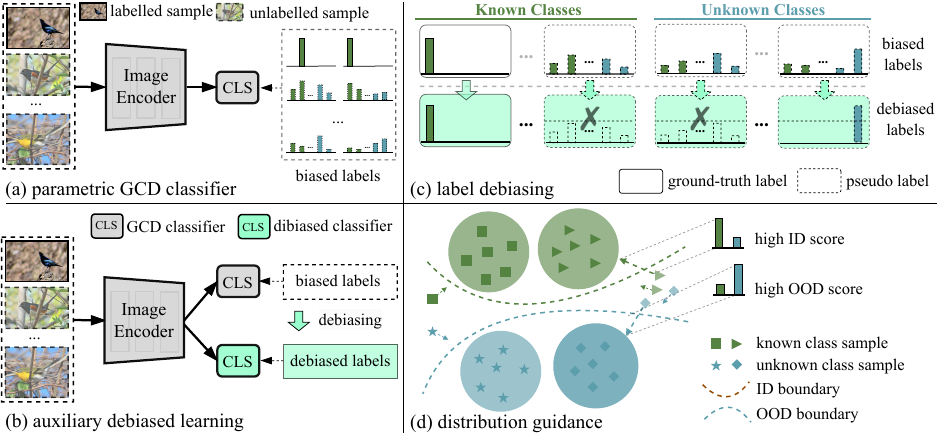}
\vspace{-5mm}
\caption{
(a) The parametric GCD classifier~\cite{wen2023parametric} is trained on labelled and unlabelled images using ground-truth hard labels and soft labels, respectively. 
(b) The auxiliary debiased learning: training another classifier using debiased labels. 
(c) The process of label debiasing: keep the hard labels unchanged and transform soft labels to one-hot hard labels; samples that do not meet the threshold are removed. 
(d) The illustration of distribution guidance: if a sample receives a high in-distribution/out-of-distribution score, its weight in GCD training will be increased accordingly.
}
\label{fig:intro}
\end{figure}

In GCD, there exists an inherent label bias between known and unknown classes due to the absence of ground-truth labels for the latter. 
This label bias has the potential to cause the model to inadvertently develop a decision rule making confident predictions that inclined to known classes. Similar problem has been identified in the area of long-tailed recognition~\cite{tang2020long,yang2022survey}. Besides, in other fields such as object classification~\cite{choi2019can,bahng2020learning,geirhosimagenet}, it is widely known that model performance suffers from task-specific bias. 
State-of-the-art parametric classifier methods in GCD, such as those proposed by~\cite{wen2023parametric,zhao2023learning,vaze2023no}, leverage the self-distillation~\cite{caron2021emerging} mechanism based on soft labels generated from the model's predictions of another image view. While these methods have shown promising results, they still rely on biased labels for training (as shown in Fig.~\ref{fig:intro}~(a)). The issue of label bias remains an unattended problem in the realm of GCD. 
Additionally, existing approaches uniformly handle all unlabelled samples without explicitly accounting for their different certainty, which may introduce noise to the model training due to unreliable samples. 
Moreover, they do not explicitly address semantic shifts, especially in a scenario like GCD involving both known and unknown classes within unlabelled data. Notably, these concerns have been demonstrated to provide significant advantages in related tasks, such as open-world semi-supervised learning~\cite{cao2022open}. In this area, OpenCon~\cite{sun2022opencon} has attempted to identify novel samples based on their proximity to known prototypes. However, its performance is heavily contingent on predefined distance thresholds, ultimately yielding suboptimal accuracy. 

We propose DebGCD, a novel framework designed to tackle the challenges of GCD. DebGCD introduces \underline{Deb}iased learning with distribution guidance for \underline{GCD}, incorporating several innovative techniques specifically tailored for this task. 
Firstly, we introduce a novel auxiliary debiased learning paradigm for GCD (as shown in Fig.~\ref{fig:intro}(b) and (c)). This method entails training an auxiliary debiased classifier in the same feature space as the GCD classifier. Unlike the GCD classifier, both labelled and unlabelled data are trained using one-hot hard labels to prevent label bias between known and unknown classes. 
Secondly, to discern the semantic distribution of unlabelled samples, we propose to learn a semantic distribution detector in a decoupled normalized feature space, which we empirically find it enhance the learning effect of GCD implicitly. 
Furthermore, we propose to measure the certainty of a sample based on its semantic distribution detection score. This certainty score then enables the gradual inclusion of unlabelled samples from both known and unknown classes during training, allowing the auxiliary debiased learning to function in a curriculum learning approach (as shown in Fig.~\ref{fig:intro}(d)), thus further enhancing its performance. 
We develop our framework upon the parametric baseline~\cite{wen2023parametric}. 
By effectively incorporating these components into a unified framework, DebGCD can be trained end-to-end in a single stage while not introducing any additional computational burden during inference. 
Despite its simplicity, DebGCD attains unparalleled performance on the public GCD datasets, including the generic classification datasets CIFAR-10~\cite{krizhevsky2009learning}, CIFAR-100~\cite{krizhevsky2009learning}, and ImageNet~\cite{deng2009imagenet}, as well as the fine-grained SSB~\cite{vaze2021open} benchmark.

We make the following key contributions in this work: 
(1) We propose DebGCD, a novel framework that addresses the challenging GCD task by considering both label bias and semantic shift, marking the first exploration of these aspects for the challenging GCD task. 
(2) Within DebGCD, we propose a novel auxiliary debiased learning paradigm to optimize the clustering feature space, in conjunction with the distribution shift detector in a distinct feature space. They work tightly to enhance the model's discovery capabilities. 
(3) We introduce a curriculum learning mechanism that steers the debiased learning process using a distribution certainty score, effectively mitigating the negative impact of uncertain samples. 
(4) Through extensive experimentation on public GCD benchmarks, DebGCD consistently demonstrates its effectiveness and achieves superior performance.

\section{Related Work}
\label{sec:formatting}

\noindent\textbf{Category Discovery.} 
This task is initially studied as novel category discovery (NCD)~\cite{han2019learning}, aiming to discover categories from unlabelled data consisting of samples from novel categories, by transferring the knowledge from the labelled categories. Many methods have been proposed to tackle NCD, such as~\cite{han2019learning,han2020automatically,han2021autonovel,fini2021unified,zhao2021novel,joseph2022novel}. 
\cite{vaze2022generalized} extends NCD to a more relaxed task, generalized category discovery (GCD), wherein unlabelled datasets encompass both known and unknown categories. A baseline method is presented for this task, incorporating self-supervised representation learning and semi-supervised $k$-means clustering, and extending popular NCD methods such as RankStats~\cite{han2020automatically} and UNO~\cite{fini2021unified} to GCD. 
CiPR~\cite{hao2023cipr} proposes to bootstrap the representation by leveraging cross-instance positive relations in the partially labelled data for contrastive learning. 
\cite{cao2022open} addresses a similar problem to GCD from the perspective of semi-supervised learning. 
SimGCD~\cite{wen2023parametric} introduces a strong parametric baseline achieving promising performance improvements. 
In~\cite{vaze2023no}, a new dataset is introduced to illustrate the limitations of unsupervised clustering in GCD. 
To address these limitations, a method based on the `mean-teachers' approach is proposed. 
In~\cite{rastegar2023learn}, a category coding approach is introduced, considering category prediction as the outcome of an optimization problem. 
Recently, SPTNet~\cite{wang2024sptnet} is proposed to consider the spatial property of images and presents a spatial prompt tuning method, enabling the model to better focus on object parts for knowledge transfer. 
Moreover, an increasing number of efforts are focused on addressing category discovery from diverse perspectives. For example,~\cite{jia2021joint} concentrates on multi-modal category discovery, whereas~\cite{zhang2022grow},~\cite{ma2024happy}, and~\cite{cendra2024promptccd} investigate continual category discovery. Additionally,~\cite{pu2024federated} explores federated category discovery. Furthermore,~\cite{wang2024hilo} studies category discovery in the presence of domain shifts.

\noindent\textbf{Debiased Learning.}
The issue of bias in data and the susceptibility of machine learning algorithms to such bias have been widely recognized as crucial challenges across diverse tasks. Numerous methodologies have been developed to address and alleviate biases inherent in training datasets or tasks. 
The studies by \cite{ponce2006toward,torralba2011unbiased} elucidate that many training sets impose regularity conditions that are impractical in real-world settings, leading to machine learning models trained on such data failing to generalize in the absence of these conditions. Furthermore, recent research by \cite{hendrycks2021natural,xiao2020noise,li2020shape} demonstrate biases in state-of-the-art object recognition models towards specific backgrounds or textures associated with object classes. Additionally, \cite{sagawa2020investigation} investigate the vulnerability of overparametrized models to spurious correlations, resulting in elevated test errors for minority groups. 
Notably, large language models also exhibit biased predictions towards certain genders or races, as indicated by \cite{cheng2021fairfil}. Furthermore, the severity of biased predictions and fairness concerns related to deployed models are extensively explored across various tasks~\cite{zemel2013learning,noble2018algorithms,bolukbasi2016man}. 
In this paper, we examine the inherent \textit{label bias} in GCD, representing the initial exploration of this issue.

\noindent\textbf{Out-of-distribution Detection.} 
In the realm of out-of-distribution (OOD) detection, the objective is to identify samples or data points that originate from a distribution distinct from the one on which the model was trained, encompassing both semantic and domain distributions~\cite{yang2021generalized,wang2024dissecting}. The simplest method in this area involves utilizing the predicted softmax class probability to detect OOD samples~\cite{hendrycks2016baseline}. 
ODIN~\cite{liang2017enhancing} further enhances this approach by introducing temperature scaling and input pre-processing. Additionally, \cite{bendale2016towards} proposes an alternative approach by calculating the score for an unknown class using a weighted average of all other classes. OOD detection has been applied in various open-set tasks, such as open-set semi-supervised learning~\cite{yu2020multi} and universal domain adaptation~\cite{saito2021ovanet}, where it is utilized to select in-distribution data during training. 
In contrast, our focus lies in the exploration of semantic shift detection considering the specific challenges of GCD. 
OpenCon~\cite{sun2022opencon} has attempted to explore the semantic shift for open-world semi-supervised learning. 
However, its reliance on a predefined distance threshold to rigidly distinguish inliers and outliers leads to suboptimal accuracy. 
In contrast, our method takes a distinct approach by avoiding a rigid separation. 
We subtly utilize the predicted OOD score by our model as a guiding factor for debiased learning, further enabling a curriculum learning scheme.

\section{Preliminaries}
\label{sec:pre}

\subsection{Problem Statement}
Generalized category discovery (GCD) aims to learn a model that can not only correctly classify the unlabelled samples of known categories but also cluster those of unknown categories. 
Given an unlabelled dataset $\mathcal{D}_u = \{(\bm{x}^{u}_{i}, y^{u}_{i})\} \in \mathcal{X} \times \mathcal{Y}_u$ and a labelled dataset $\mathcal{D}_l = \{(\bm{x}^{l}_{i}, y^{l}_{i})\} \in \mathcal{X} \times \mathcal{Y}_l$, where $\mathcal{Y}_u$ and $\mathcal{Y}_l$ are their label sets respectively. 
The unlabelled dataset contains samples from both known and unknown categories, \textit{i.e.}, $\mathcal{Y}_l \subset \mathcal{Y}_u$.
The number of labelled categories is $M=|\mathcal{Y}_l|$. We  assume the number of categories $K=|\mathcal{Y}_l \cup \mathcal{Y}_u|$ to be known following previous works~\cite{han2021autonovel,wen2023parametric,vaze2023no}. When it is unknown, methods like~\cite{han2019learning,vaze2022generalized} can be applied to provide a reliable estimation. 

\subsection{Baseline}
\cite{wen2023parametric} introduces a robust parametric GCD baseline, which has been widely adopted in the field ever since~\cite{vaze2023no,wang2024sptnet}. It employs a parametric classifier, implemented in a self-distillation manner~\cite{caron2021emerging}. 
The classifier is randomly initialized with $K$ normalized category prototypes $\mathcal{C}=\{\bm{c}_1,...,\bm{c}_K\}$. For the randomly augmented view of an image $\bm{x}_i$ and its corresponding normalized hidden feature vector $\bm{h}_i=\phi(\bm{x}_i)/||\phi(\bm{x}_i)||$, the output probability for the $k$th category is given by: 
\begin{equation}
    {\bm{p}_i}^{(k)} = \frac{\exp(\bm{h}_i\cdot\bm{c}_k/\tau_s)}{\sum\nolimits_{j=1}^K \exp(\bm{h}_i\cdot\bm{c}_j/\tau_s)},
\end{equation}
where $\tau_s$ is the scaling temperature for this \textit{`student'} view. The soft label $\bm{q}_i$ is produced by the \textit{`teacher'} view with a sharper temperature $\tau_t$ using another augmented view in the same fashion. 
The self-distillation loss of the two views is then simply calculated following the cross-entropy loss $\ell_{ce}(\bm{q}',\bm{p})=-\sum\nolimits_{j=1}^K \bm{q}'^{(j)}\text{log}~\bm{p}^{(j)}$. 
Given a mini-batch $\mathcal{B}$ containing both labelled samples $\mathcal{B}_l$ and unlabelled images $\mathcal{B}_u$, the self-distillation loss is calculated using all the samples in the mini-batch:
\begin{equation}
    \mathcal{L}_{cls}^{u}=\frac{1}{|\mathcal{B}|}\sum_{i\in \mathcal{B}} \ell_{ce}(\bm{q}'_i,\bm{p}_i)-\xi H(\overline{\bm{p}}), 
\end{equation}
where $\overline{\bm{p}}=\frac{1}{2|\mathcal{B}|}\sum\nolimits_{i\in \mathcal{B}}(\bm{p}_i+\bm{p}'_i)$ denotes the mean prediction within a batch and its entropy $H(\overline{\bm{p}})=-\sum\nolimits_{j=1}^K \overline{\bm{p}}^{(j)}\text{log}~\overline{\bm{p}}^{(j)}$ weighted by $\xi$. 
For the labelled samples, the supervised classification loss is written as $\mathcal{L}_{cls}^s=\frac{1}{|\mathcal{B}_l|}\sum\nolimits_{i\in \mathcal{B}_l}\ell_{ce}(\bm{p}_i,\bm{y}_i)$, where $\bm{y}_i$ represents the one-hot vector corresponding to the ground-truth label $y_i$. 
The whole classification objective is $\mathcal{L}_{cls}=(1-\lambda_b^{gcd})\mathcal{L}_{cls}^u+\lambda_b^{gcd}\mathcal{L}_{cls}^s$. Combining with the representation learning loss $\mathcal{L}_{rep}$ adopted from~\cite{vaze2022generalized}, the overall training objective becomes:
\begin{equation}
    \mathcal{L}_{gcd} = \mathcal{L}_{cls} + \mathcal{L}_{rep}.
\end{equation}
Through training with $\mathcal{L}_{gcd}$ on both $\mathcal{D}_l$ and $\mathcal{D}_u$, the classifier can directly predict the labels for unlabelled samples after training.

\begin{figure*}[t]
\centering
\includegraphics[width = \textwidth]{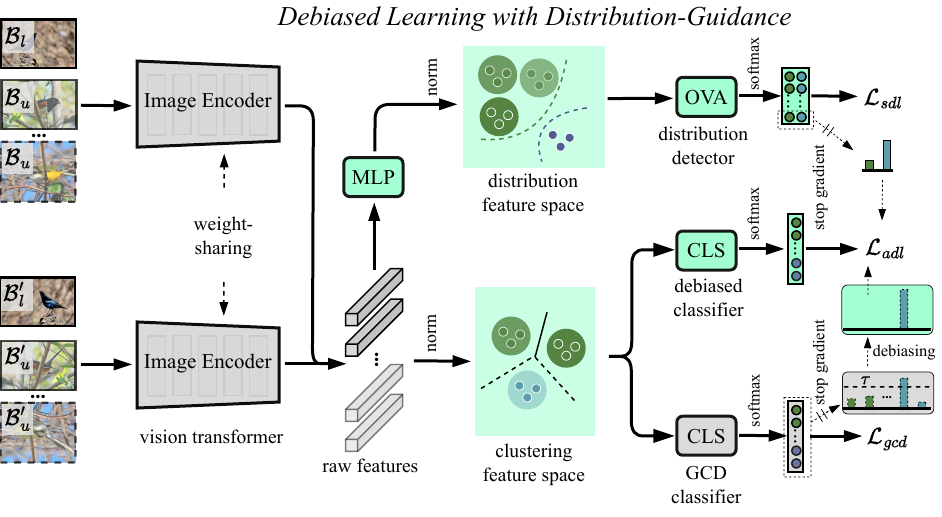}
\vspace{-8mm}
\caption{
Overview of the DebGCD framework. 
In the upper branch, raw features are transformed using an MLP and then normalized. These normalized features are used for semantic distribution learning with a one-vs-all classifier. In the lower branch, a GCD classifier is trained on the normalized raw features. The predictions from both branches are combined to train the debiased classifier. As DebGCD aligns with prior work in representation learning, it's not explicitly depicted here.
}
\label{fig:method}
\end{figure*}

\section{Debiased Learning with Distribution-Guidance for GCD}
In this section, we present our debiased Learning with distribution-guidance framework for GCD (see Fig.~\ref{fig:method}). 
First, in Sec.~\ref{sec:ood}, we present the semantic distribution learning on the GCD task. Next, in Sec.~\ref{sec:st}, we demonstrate the training paradigm of the debiased classifier. 
Finally, we describe the joint training and inference process of our full framework in Sec.~\ref{sec:jt}.

\subsection{Learning Semantic Distribution}
\label{sec:ood}
OOD detection methods have been employed in tasks like universal domain adaptation~\cite{saito2021ovanet} and open-set semi-supervised learning~\cite{yu2020multi}, obtaining improved performance. 
In these tasks, the identified OOD samples are treated as a single \textit{background} class to avoid affecting the recognition of unlabelled samples from the labelled classes, and the distribution shifts can be of any type. 
In GCD, we are particularly interested in identifying the semantic shifts. 
The instances from the labelled classes are considered in-distribution (ID) samples, while the instances from the novel classes are considered OOD samples. 
However, the potential of effectively introducing OOD techniques for GCD remains under-explored. 
An intuitive approach for OOD detection is to examine the class prediction probabilities. 
Generally, the maximum softmax or logit score of a closed-set classifier can serve as a good indicator of OOD~\cite{vaze2021open,wang2024dissecting}. 
However, this is not suitable for the common GCD classifier, which contains an mean entropy regularization term in the loss function to prevent biased predictions. 
Nevertheless, we find that it also results in the classifier's predictions on known categories being less confident, thereby degrading the OOD detection performance. 
Moreover, these OOD methods need to carefuly select a threshold~\cite{geng2020recent} for rejecting ``unknown'' samples, which relies on validation or a pre-defined ratio of ``unknown'' samples, making them impractical for the GCD due to the absence of such validation samples. 
One-vs-all (OVA) classifier~\cite{saito2021ovanet}, which has consistently shown promise in the literature~\cite{saito2021openmatch,fan2023ssb,li2023iomatch}, can be a more suitable option. 
Moreover, in the context of OOD, the objective is not to differentiate between multiple distinct unknown categories, as we do in GCD; rather, we aim to distinguish all unknown classes from the known classes, effectively framing this as a binary classification problem. This calls for the need of a different feature space that is better suited for this task. 
Therefore,  as depicted in Fig.~\ref{fig:method}, we introduce an additional multi-layer perceptron (MLP) projection network $\rho_s$, to project raw features into another embedding space, followed by $\ell_2$-normalization to attain the embedding space for distribution discrimination. 
Different from the prior works applying OOD in the magnitude-aware feature space for other tasks~\cite{yu2020multi,saito2021openmatch,li2023iomatch}, we empirically found that the $\ell_2$-normalized feature space aligns more seamlessly with the DINO pre-trained weights in GCD. 
Subsequently, we devise $M$ $\ell_2$-normalized binary classifiers, denoted as $\chi = \{\chi_1,\chi_2,...,\chi_M\}$, for semantic OOD detection in GCD. 

Given an augmented image view $\bm{x}_i$, its $\ell_2$-normalized feature in the semantic distribution feature space is $\bm{f}_i = \rho_s(\phi(\bm{x}_i))/||\rho_s(\phi(\bm{x}_i))||$. Subsequently, the output of the $k$-th binary classifier is $\bm{o}_{i,k} = \text{softmax}(\chi_k(\bm{f}_i))$, 
where $\bm{o}_{i,k}=(o^{+}_{i,k}, o^{-}_{i,k})$ and $o^{+}_{i,k}+o^{-}_{i,k}=1$. 
For labelled samples, a multi-binary cross-entropy loss with a hard-negative sampling strategy~\cite{saito2021openmatch} is employed:
\begin{equation}
    \mathcal{L}_{sdl}^{s} = \frac{1}{|\mathcal{B}_l|}\sum_{i\in \mathcal{B}_l}(-\log(o^{+}_{i,y_i})-\min_{k\neq y_i} \log(o^{-}_{i,k})),
\end{equation}
where $y_i$ represents the ground-truth category label of the sample $\bm{x}_i$. For unlabelled samples, an entropy minimization technique~\cite{saito2021ovanet} is applied to improve low-density separation:
\begin{equation}
    \mathcal{L}_{sdl}^u = -\frac{1}{\mathcal{B}_u}\sum_{i\in \mathcal{B}_u}\sum_{j=1}^{M}(o^{+}_{i,j}\log(o^{+}_{i,j}) + o^{-}_{i,k}\log(o^{-}_{i,k})),
\end{equation}
where $\mathcal{B}_u$ denotes the unlabelled subset in current mini-batch. The loss function for the semantic distribution learning is defined as:
\begin{equation}
    \mathcal{L}_{sdl} = \mathcal{L}_{sdl}^s + \mathcal{L}_{sdl}^u.
\end{equation}
By optimizing $\mathcal{L}_{sdl}$, our detector distinctly segregates the feature distributions between known and unknown categories. 
Additionally, it generates a predicted score based on the maximum output from all $M$ binary classifiers, denoted as:
\begin{equation}
    s_i = o^{-}_{i, y_p}, y_p = \arg\max_{j}o^{+}_{i,j}.
\end{equation}
This score will serve as a crucial cue for the debiased learning to be introduced next.

\subsection{Auxiliary Debiased Learning}
\label{sec:st}
As depicted in Fig.~\ref{fig:method}, the raw features are normalized to the clustering feature space in the lower branch, wherein novel categories are discovered. 
In order to minimize the unintended negative impact of biased labels while maintaining the basic probability constraints~\cite{assran2022masked} and consistency regularization~\cite{caron2021emerging} in the GCD classifier, we propose an auxiliary debiased learning mechanism. 
Specifically, a parallel debiased classifier $\psi_s$ initialized with $K$ normalized prototypes $\mathcal{C}^{a}=\{\bm{c}^{a}_1,...,\bm{c}^{a}_K\}$, is trained in the same embedding space using debiased labels. 
Note that in our experiment, we only finetune the last two transformer blocks of the DINO~\cite{caron2021emerging} pre-trained ViT backbone. 
The $k$-th softmax score of sample $\bm{x}_i$ is given by:
\begin{equation}
    {\bm{p}_i^{a}}^{(k)} = \frac{\exp(\bm{h}_i \cdot \bm{c}^{a}_k/\tau_{a})}{\sum\nolimits_{j=1}^K \exp(\bm{h}_i \cdot \bm{c}^{a}_j/\tau_{a})},
\end{equation}
where $\tau_{a}$ is the scaling temperature. The maximum classification score has demonstrated promising performance in several semi-supervised learning methods and we find it also a good indicator of sample quality in the context of GCD task. For an augmented view $\bm{x}_i$ and its GCD classifier prediction $\bm{p}_i$, a debiasing threshold $\tau$ is set on the $\max(\bm{p}_i)$, with only samples surpassing $\tau$ being utilized to train the debiased classifier, expressed as $\mathbbm{1}(\max(\bm{p}_i)>\tau)$. Additionally, given that the semantic distribution detector and the GCD classifier are learned in different feature spaces and paradigms, it is essential to ensure the alignment of their predictions. 
Consequently, we introduce a function to indicate the task consistency of these two tasks, defined as:
\begin{equation}
    \mathcal{F}(\hat{y_i}, s_i) = \mathbbm{1}(\hat{y_i} \in \mathcal{Y}_u \land s_i>0.5) \lor \mathbbm{1}(\hat{y_i} \in \mathcal{Y}_l \land s_i<0.5),
\end{equation}
where $\hat{y_i} = \arg\max(\bm{p}_i)$ represents the predicted category index by the GCD classifier, and $\hat{\bm{y}_i}$ denotes its corresponding one-hot vector. This function aims to selectively filter out samples with identical distribution predictions across the two tasks.

Furthermore, as previously stated, given the inclusion of both known (in-distribution) and unknown (out-of-distribution) samples in the unlabelled data, it is imperative to devise a learning strategy based on semantic distribution information. 
With the training progresses, the semantic OOD scores gradually approach the two extremes (\textit{i.e.}, $0$ and $1$). 
The score of the unknown class sample steadily increases to $1$, while the score of the known class gradually decreases to $0$. Prior techniques~\cite{saito2021openmatch,li2023iomatch} simply employ a threshold to determine whether the sample belongs to the known or unknown. 
Such a na\"ive method is unreliable and may introduce many noises to the model training for GCD.  
In our approach, we prioritize samples with distinct distributions for self-training, aligning with the principles of curriculum learning. To establish a consistent metric for assessing sample discriminability, we introduce a normalized distribution certainty score:
\begin{equation}
    d_i = |2\times s_i - 1|,
\end{equation}
which approaches the value $0$ for ambiguous samples and the value $1$ for certain samples. This score, to a certain extent, indicates the learning status of samples and can serve as a crucial cue for our debiased classifier. 
Therefore, the auxiliary debiased learning loss for unlabelled samples is written as:
\begin{equation}
    \mathcal{L}_{adl}^u = \frac{1}{\mathcal{B}_u}\sum_{i\in \mathcal{B}_u}\mathbbm{1}(\max(\bm{p_i})>\tau)\times \mathcal{F}(\hat{y_i}, s_i)\times d_i\times \ell_{ce}({\bm{p}_i^{a}},\hat{\bm{y}_i}).
\end{equation}
In this manner, the training of the debiased classifier transforms into a curriculum learning process, where easily identifiable samples that are clearly semantic in-distribution or out-of-distribution are given higher priority for learning. 
Moreover, our debiased classifier also retains the prior knowledge from the labelled data. For the labelled samples, it's is simply trained with the cross-entropy loss: 
\begin{equation}
    \mathcal{L}_{adl}^s = \frac{1}{\mathcal{B}_l}\sum_{i\in \mathcal{B}_l}\ell_{ce}({\bm{p}_i^{a}},\bm{y}_i).
\end{equation}
Finally, the overall training loss for the debiased classifier is:
\begin{equation}
    \mathcal{L}_{adl} = \mathcal{L}_{adl}^s + \mathcal{L}_{adl}^u.
\end{equation}
In this manner, all the samples are trained using one-hot hard labels, irrespective of their belongings to known or unknown categories. Operating within the same feature space, our debiased classifier collaborates closely with the GCD classifier, thereby facilitating the joint optimization of the clustering feature space.

\begin{algorithm}[t]
\caption{End-to-end Training Algorithm for DebGCD.}
\label{alg:dist}
\textbf{Input}: Set of labelled data $\mathcal{D}_l = \{(\bm{x}^{l}_{i}, y^{l}_{i})\}$, set of unlabelled data $\mathcal{D}_u = \{(\bm{x}^{u}_{i}, y^{u}_{i})\}$. Data augmentation function $\mathcal{A}$. Model parameters $w$, learning rate $\eta$, epoch $E_{max}$, iteration $I_{max}$, trade-off parameters, $\lambda_{sdl}$, $\lambda_{adl}$;

\textbf{for} $Epoch = 1~~to~~E_{max}$ \textbf{do}

~~~~\textbf{for} $Iteration = 1~~to~~I_{max}$ \textbf{do}

~~~~~~~~\textbf{Sample} labelled data $\mathcal{B}_l$, unlabelled data $\mathcal{B}_u$;~$i\in \mathcal{B}_u$ 

~~~~~~~~\textbf{Compute} model prediction $\bm{p}_i$, $\bm{p}_i^{a}$, $s_i$; loss function $\mathcal{L}_{gcd}$, $\mathcal{L}_{sdl}$
~~~~~~~~~~~~~~~~~~~~~~~~~~~~~~~~//~Eq.3,6,8

~~~~~~~~\textbf{Compute} debiased label $\hat{y_i}$; task consistency $\mathcal{F}(\hat{y_i}, s_i)$~~~~~~~~~~~~~~~~~~~~~~~~~~~~~~~~~~~~~~~~~~~~~~~~~~~~~~~//~Eq.9

~~~~~~~~\textbf{Compute} loss function $\mathcal{L}_{adl}^s$, $\mathcal{L}_{adl}^u$, $\mathcal{L}_{adl}$~~~~~~~~~~~~~~~~~~~~~~~~~~~~~~~~~~~~~~~~~~~~~~~~~~~~~~~~~~~~~~~~//~Eq.11,12,13

~~~~~~~~\textbf{Compute} loss function $\mathcal{L}_{all}=\mathcal{L}_{gcd}+\lambda_{sdl}\mathcal{L}_{sdl}+\lambda_{adl}\mathcal{L}_{adl}$

~~~~~~~~\textbf{Update} model parameters $w = w - \eta \bigtriangledown_w\mathcal{L}_{all}$

~~~~\textbf{end}

\textbf{end}

\textbf{Output}: Model parameter $w$.
\end{algorithm}

\subsection{Learning and Inference Framework}
\label{sec:jt}

Based on the baseline GCD classifier, our framework is designed to be trained in a multi-task manner. Different from previous approaches in the open-set literature~\cite{yu2020multi}, our DebGCD framework employs a \textit{one-stage} training process, eliminating the necessity for task-specific warm-up phases. Consequently, the three tasks can be jointly trained end-to-end with the overall loss: 
\begin{equation}
\mathcal{L}_{all}=\mathcal{L}_{gcd}+\lambda_{sdl}\mathcal{L}_{sdl}+\lambda_{adl}\mathcal{L}_{adl},
\end{equation}
where $\lambda_{sdl}$ and $\lambda_{adl}$ denote the loss weights for the semantic distribution detector and debiased classifier, respectively. 
The complete training pipeline of the framework is illustrated in Algorithm~\ref{alg:dist}.

Throughout the joint training process, the three branches are collectively optimized in an end-to-end manner. 
During inference, only the GCD classifier is retained. 
This indicates that our method does not impose any additional computational overhead compared to the baseline approach during inference, further emphasizing its simplicity and efficiency.

\section{Experiments}
\label{sec:exp}
In this section, we present a comprehensive evaluation of the proposed DebGCD framework and further perform meticulous ablation studies to showcase the effectiveness of its individual components. More results and analysis can be found in the Appendix.

\subsection{Experimental Setup}
\noindent\textbf{Datasets.}
We conduct a comprehensive evaluation of our method across diverse benchmarks, encompassing the generic image recognition benchmark (CIFAR-10/100~\cite{krizhevsky2009learning}, ImageNet-100~\cite{deng2009imagenet}), the Semantic Shift Benchmark (SSB)~\cite{vaze2022semantic} comprising fine-grained  datasets CUB~\cite{wah2011caltech}, Stanford Cars~\cite{krause20133d}, and FGVC-Aircraft~\cite{maji2013fine}, along with the challenging ImageNet-1K~\cite{deng2009imagenet}. For each dataset, we adhere to the data split scheme detailed in~\cite{vaze2022generalized}. The method involves sampling a subset of all classes as the known (`Old') classes $\mathcal{Y}_l$. Subsequently, $50\%$ of the images from these known classes are utilized to construct $\mathcal{D}_l$, while the remaining images are designated as the unlabelled data $\mathcal{D}_u$. The statistics can be seen in Tab.~\ref{tab:datasets}.

\begin{wraptable}{R}{0.6\textwidth}
\centering
\caption{Overview of dataset, including the classes in the labelled and unlabelled sets ($|\mathcal{Y}_l|$, $|\mathcal{Y}_u|$) and counts of images ($|\mathcal{D}_l|$, $|\mathcal{D}_u|$). `FG' denotes fine-grained.}
\setlength{\tabcolsep}{1mm}{
\resizebox{0.6\textwidth}{!}{
\begin{tabular}{lccccc}
    \toprule
    Dataset &FG &$|\mathcal{D}_l|$&$|\mathcal{Y}_l|$&$|\mathcal{D}_u|$ &$|\mathcal{Y}_u|$\\
    \midrule
    CIFAR-10~\cite{krizhevsky2009learning} &\ding{55} & 12.5K & 5 & 37.5K & 10 \\
    CIFAR-100~\cite{krizhevsky2009learning} &\ding{55} & 20.0K & 80 & 30.0K & 100 \\
    ImageNet-100~\cite{deng2009imagenet} &\ding{55} & 31.9K & 50 & 95.3K & 100 \\
    CUB~\cite{wah2011caltech} &\ding{51} & 1.5K & 100 & 4.5K & 200 \\
    Stanford Cars~\cite{krause20133d} &\ding{51} & 2.0K & 98 & 6.1K & 196 \\
    FGVC-Aircraft~\cite{maji2013fine} & \ding{51} &1.7K & 50 & 5.0K & 100 \\
    ImageNet-1K~\cite{deng2009imagenet} &\ding{55} & 321K & 500 & 960K & 1000 \\
    \bottomrule
\end{tabular}
}
}
\label{tab:datasets}
\end{wraptable}

\noindent\textbf{Evaluation metrics.}
We assess the GCD performance using the clustering accuracy ($\text{ACC}$) in accordance with established conventions~\cite{vaze2022generalized}. For evaluation, the $\text{ACC}$ on $\mathcal{D}_l$ is computed as follows, given the ground truth $y_i$ and the predicted labels $\hat{y}_i$:
\begin{equation}
    \text{ACC}=\frac{1}{|\mathcal{D}_u|}\sum\limits_{i=1}^{|\mathcal{D}_u|}\mathbbm{1}(y_i=h(\hat{y}_i)), 
\end{equation}
where $h$ represents the optimal permutation that aligns the predicted cluster assignments with the ground-truth class labels. $\text{ACC}$ for `All' classes, `Old' classes and `New' classes are reported for comprehensive assessment.

\noindent\textbf{Implementation details.}
Following previous attempts in GCD~\cite{vaze2022generalized,wen2023parametric}, our model is structured with a ViT-B/16~\cite{dosovitskiy2020image} backbone pre-trained using DINO~\cite{caron2021emerging}, and the feature space centers around the $768$-dimensional classification token. The projection networks for representation learning and semantic distribution detection comprise three-layer and five-layer MLPs, respectively. The model is trained with a batch size of 128, initiating with an initial learning rate of $10^{-1}$ which decays to $10^{-4}$ using a cosine schedule over $200$ epochs. Notably, the loss weights $\lambda_{sdl}$ and $\lambda_{adl}$ are set to $0.01$ and $1.0$, while the loss balancing weight $\lambda_b^{gcd}$ is assigned to $0.35$ following~\cite{wen2023parametric}. Regarding the temperature parameters, the initial temperature $\tau_t$ is established at $0.07$, subsequently warmed up to $0.04$ employing a cosine schedule during the first 30 epochs, whereas the other temperatures are set to $0.1$.

\subsection{Benchmark Results}
We present benchmark results of our method and compare it with state-of-the-art techniques in generalized category discovery (including ORCA~\cite{cao2022open}, GCD~\cite{vaze2022generalized}, XCon~\cite{fei2022xcon}, OpenCon~\cite{sun2022opencon},  PromptCAL~\cite{zhang2023promptcal}, DCCL~\cite{pu2023dynamic}, GPC~\cite{zhao2023learning}, CiPR~\cite{hao2023cipr}, SimGCD~\cite{wen2023parametric}, $\mu$GCD~\cite{vaze2023no}, InfoSieve~\cite{rastegar2023learn}, and SPTNet~\cite{wang2024sptnet}), as well as robust baselines derived from novel category discovery (RankStats+~\cite{han2021autonovel}, UNO+~\cite{fini2021unified}, and $k$-means~\cite{macqueen1967some}). All methods are based on the DINO~\cite{caron2021emerging} pre-trained backbone. 
This comparative evaluation encompasses performance on the fine-grained SSB benchmark~\cite{vaze2022semantic} and generic image recognition datasets~\cite{krizhevsky2009learning, deng2009imagenet}, as shown in Tab.~\ref{tab:ssb} and Tab.~\ref{tab:generic}.

\noindent\textbf{Results on SSB.}
As shown in Tab.~\ref{tab:ssb}, DebGCD demonstrates superior performance across the three datasets, achieving an average ACC of $64.4$ on `All' categories, surpassing the second-best SPTNet~\cite{wang2024sptnet} by $3\%$. It maintains the best on both Stanford Cars and FGVC-Aircraft dataset, while ranking second on CUB, where it is outperformed only by InfoSieve~\cite{rastegar2023learn}, a hierarchical encoding method specifically designed for fine-grained GCD. In contrast, DebGCD aims for broader improvements across both generic and fine-grained datasets. 
These results reveal DebGCD's exceptional ability to uncover new categories, while also showcasing remarkable performance in recognizing known categories. 

\begin{table*}[t]
\centering
\caption{Comparison of state-of-the-art GCD methods on SSB~\cite{vaze2022semantic} benchmark. Results are reported in ACC across the `All', `Old' and `New' categories.}
\setlength{\tabcolsep}{2mm}{
\resizebox{0.9\textwidth}{!}{
\begin{tabular}{l>{\columncolor{my_blue}}ccc>{\columncolor{my_blue}}ccc>{\columncolor{my_blue}}ccc>{\columncolor{my_blue}}ccc>{\columncolor{my_blue}}c}
    \toprule
     &\multicolumn{3}{c}{CUB}&\multicolumn{3}{c}{Stanford Cars}&\multicolumn{3}{c}{FGVC-Aircraft}&\multicolumn{1}{c}{Average}\\
    \cmidrule(lr{1em}){2-4} \cmidrule(lr{1em}){5-7} \cmidrule(lr{1em}){8-10} 
 Method&All&Old&New&All&Old&New&All&Old&New&All\\ \hline
    $k$-means~\cite{macqueen1967some}&34.3&38.9&32.1 &12.8&10.6&13.8 &16.0&14.4&16.8 &21.1\\
    RankStats+~\cite{han2021autonovel}&33.3&51.6&24.2 &28.3&61.8&12.1 &26.9&36.4&22.2 &29.5\\
    UNO+~\cite{fini2021unified}&35.1&49.0&28.1 &35.5&70.5&18.6 &40.3&56.4&32.2 &37.0\\
    ORCA~\cite{cao2022open}&35.3&45.6&30.2 &23.5&50.1&10.7 &22.0&31.8&17.1 &26.9\\
    GCD~\cite{vaze2022generalized}&51.3&56.6&48.7 &39.0&57.6&29.9 &45.0&41.1&46.9 &45.1\\
    XCon~\cite{fei2022xcon}&52.1&54.3&51.0 &40.5&58.8&31.7 &47.7&44.4&49.4 &46.8\\
    OpenCon~\cite{sun2022opencon}&54.7&63.8&54.7 &49.1&78.6&32.7 &-&-&- &-\\
    PromptCAL~\cite{zhang2023promptcal}&62.9&64.4&62.1 &50.2&70.1&40.6 &52.2&52.2&52.3 &55.1\\
    DCCL~\cite{pu2023dynamic}&63.5&60.8&64.9 &43.1&55.7&36.2 &-&-&- &-\\
    GPC~\cite{zhao2023learning}&52.0&55.5&47.5 &38.2&58.9&27.4 &43.3&40.7&44.8 &44.5\\
    SimGCD~\cite{wen2023parametric}&60.3&65.6&57.7 &53.8&71.9&45.0 &54.2&59.1&51.8 &56.1\\
    $\mu$GCD~\cite{vaze2023no} &65.7&68.0&64.6 &56.5&68.1&\underline{50.9} &53.8&55.4&53.0 &58.7\\
    InfoSieve~\cite{rastegar2023learn}&\textbf{69.4}&\textbf{77.9}&\textbf{65.2} &55.7&74.8&46.4 &56.3&\underline{63.7}&52.5 &60.5\\
    CiPR~\cite{hao2023cipr} &57.1&58.7&55.6 &47.0&61.5&40.1 &-&-&- &-\\
    SPTNet~\cite{wang2024sptnet} &65.8&68.8&\underline{65.1} &\underline{59.0}&\underline{79.2}&49.3 &\underline{59.3}&61.8&\underline{58.1} &\underline{61.4}\\
    \hline
    \textbf{DebGCD  }&\underline{66.3}&\underline{71.8}&63.5 &\textbf{65.3}&\textbf{81.6}&\textbf{57.4} &\textbf{61.7}&\textbf{63.9}&\textbf{60.6} &\textbf{64.4}\\
    \bottomrule
\end{tabular}
}
}
\label{tab:ssb}
\end{table*}

\begin{table*}[t]
\centering
\caption{Comparison of state-of-the-art GCD methods on generic datasets. It includes CIFAR-10~\cite{krizhevsky2009learning}, CIFAR-100~\cite{krizhevsky2009learning}, ImageNet-100~\cite{deng2009imagenet}, and ImageNet-1K~\cite{deng2009imagenet} dataset.}
\setlength{\tabcolsep}{2.0mm}{
\resizebox{0.97\textwidth}{!}{
\begin{tabular}{l>{\columncolor{my_blue}}ccc>{\columncolor{my_blue}}ccc>{\columncolor{my_blue}}ccc>{\columncolor{my_blue}}ccc>{\columncolor{my_blue}}ccc}
    \toprule
     &\multicolumn{3}{c}{CIFAR-10}&\multicolumn{3}{c}{CIFAR-100}&\multicolumn{3}{c}{ImageNet-100}&\multicolumn{3}{c}{ImageNet-1K}\\
    \cmidrule(lr{1em}){2-4} \cmidrule(lr{1em}){5-7} \cmidrule(lr{1em}){8-10} \cmidrule(lr{1em}){11-13}
 Method&All&Old&New&All&Old&New&All&Old&New &All&Old&New\\ \hline
    $k$-means~\cite{macqueen1967some} &83.6&85.7&82.5 &52.0&52.2&50.8 &72.7&75.5&71.3 &-&-&-\\
    RankStats+~\cite{han2021autonovel} &46.8&19.2&60.5 &58.2&77.6&19.3 &37.1&61.6&24.8 &-&-&-\\
    UNO+~\cite{fini2021unified} &68.6&\textbf{98.3}&53.8 &69.5&80.6&47.2 &70.3&\textbf{95.0}&57.9 &-&-&-\\
    ORCA~\cite{cao2022open} &69.0&77.4&52.0 &73.5&\textbf{92.6}&63.9 &81.8&86.2&79.6 &-&-&-\\
    GCD~\cite{vaze2022generalized} &91.5&\underline{97.9}&88.2 &73.0&76.2&66.5 &74.1&89.8&66.3 &52.5&72.5&42.2\\
    XCon~\cite{fei2022xcon} &96.0&97.3&95.4 &74.2&81.2&60.3 &77.6&93.5&69.7 &-&-&-\\
    OpenCon~\cite{sun2022opencon} &-&-&- &-&-&- &84.0&93.8&81.2 &-&-&-\\
    PromptCAL~\cite{zhang2023promptcal}  &\textbf{97.9}&96.6&\underline{98.5} &81.2&84.2&75.3 &83.1&92.7&78.3 &-&-&-\\
    DCCL~\cite{pu2023dynamic} &96.3&96.5&96.9 &75.3&76.8&70.2 &80.5&90.5&76.2 &-&-&-\\
    GPC~\cite{zhao2023learning} &90.6&97.6&87.0 &75.4&\underline{84.6}&60.1 &75.3&93.4&66.7 &-&-&-\\
    SimGCD~\cite{wen2023parametric} &97.1&95.1&98.1 &80.1&81.2&77.8 &83.0&93.1&77.9 &\underline{57.1}&\underline{77.3}&\underline{46.9}\\
    InfoSieve~\cite{rastegar2023learn}&94.8&97.7&93.4 &78.3&82.2&70.5 &80.5&93.8&73.8 &-&-&-\\
    CiPR~\cite{hao2023cipr} &\underline{97.7}&97.5&97.7 &\underline{81.5}&82.4&\underline{79.7} &80.5&84.9&78.3 &-&-&- \\
    SPTNet~\cite{wang2024sptnet} &97.3&95.0&\textbf{98.6} &81.3&84.3&75.6 &\underline{85.4}&93.2&\underline{81.4} &-&-&-\\
    \hline
    \textbf{DebGCD  }&97.2&94.8&98.4 &\textbf{83.0}&\underline{84.6}&\textbf{79.9} &\textbf{85.9}&\underline{94.3}&\textbf{81.6} &\textbf{65.0}&\textbf{82.0}&\textbf{56.5}\\
    \bottomrule
\end{tabular}
}
}
\label{tab:generic}
\end{table*}

\noindent\textbf{Results on generic datasets.}
In Tab.~\ref{tab:generic}, we report results on three widely used generic datasets (CIFAR-10, CIFAR-100 and ImageNet-100) in GCD, as well as the challenging ImageNet-1K. 
Our method attains superior performance in terms of ACC across `All' categories, establishing the new state-of-the-art, except CIFAR-10, on which the performance is nearly saturated (over 97\% ACC) for our method and other most competitive methods. 
On the challenging ImageNet-1K, containing $1,000$ classes with diverse images, DebGCD also establishes the new state-of-the-art, surpassing the previous best-performing method by $7.9\%$. 
These results validate the effectiveness and robustness of our method for generalized category discovery on generic datasets.

\begin{table}[tp]
\centering
\caption{Ablations. The results regarding the different components in our framework on Stanford Cars~\cite{krause20133d}. ACC of `All', `Old' and `New' categories are listed.}
\setlength{\tabcolsep}{2mm}{
\resizebox{0.7\textwidth}{!}{
\begin{tabular}{cccccccc}
    \toprule
&\multirow{2}*{\shortstack{Debiased \\ Learning}}&\multirow{2}*{\shortstack{Auxiliary \\Classifier}}&\multirow{2}*{\shortstack{Semantic Dist. \\Learning}}&\multirow{2}*{\shortstack{Dist. \\Guidance}}&\multicolumn{3}{c}{Stanford Cars}\\
\cmidrule(lr{1em}){6-8}
&&&& &All&Old&New \\
    \midrule
    (1)&\ding{55}&\ding{55}&\ding{55}&\ding{55} &53.8&71.9&45.0 \\
    (2)&\ding{51}&\ding{55}&\ding{55}&\ding{55} &51.3&72.8&40.9 \\
    (3)&\ding{51}&\ding{51}&\ding{55}&\ding{55} &58.5&78.7&48.8 \\
    (4)&\ding{55}&\ding{55}&\ding{51}&\ding{55} &56.5&73.3&48.3 \\
    (5)&\ding{51}&\ding{51}&\ding{51}&\ding{55} &60.7&78.1&52.3 \\
    \rowcolor{my_blue} (6)&\ding{51}&\ding{51}&\ding{51}&\ding{51} &\textbf{65.3}&\textbf{81.6}&\textbf{57.4} \\
    \bottomrule
\end{tabular}
}
}
\label{tab:ablation1}
\end{table}

\subsection{Analysis}
In this section, we provide ablations regarding the key components within our framework. Besides, we study the impact of the debiasing threshold $\tau$ and labelled data. 
\begin{wraptable}{R}{0.35\textwidth}
\centering
\caption{Experimental results on distillation data by using different loss functions.}
\setlength{\tabcolsep}{3mm}{
\resizebox{0.35\textwidth}{!}{
\begin{tabular}{ccccc}
        \toprule
        \multirow{2}*{$\mathcal{L}_{adl}^s$}&\multirow{2}*{$\mathcal{L}_{adl}^u$}&\multicolumn{3}{c}{FGVC-Aircraft}\\
        \cmidrule(lr{1em}){3-5} 
        & &All&Old&New \\
        \midrule
        & &54.2&59.1&51.8 \\
        \ding{51}& &53.1&60.5&49.4 \\
        &\ding{51} &57.9&60.1&56.9 \\
        \rowcolor{my_blue} \ding{51}&\ding{51} &\textbf{61.7} &\textbf{63.9} &\textbf{60.6} \\
        \bottomrule
        \end{tabular}
}
}
\label{tab:data}
\end{wraptable}

\noindent\textbf{Framework components.}
Starting with the baseline method trained using $\mathcal{L}_{gcd}$ (Row (1)), we gradually incorporate our proposed techniques on the Stanford Cars dataset, as depicted in Tab.~\ref{tab:ablation1}. 
An intuitive approach is to apply debiased learning to the original classifier as in Row~(2). However, this still produces a biased supervision signal because it relies on the original GCD loss for that classifier. It turns out that such a na\"ive approach may even hurt the performance. 
Rows~(1) and (2) indicate that directly applying debiased learning to the GCD classifier can lead to a decrease in performance, particularly affecting novel categories. 
The introduction of an auxiliary classifier in Row~(3) demonstrates significant performance enhancements. Similarly, our semantic distribution learning alone results in a $2.7\%$ improvement across all categories in Row~(4). Row~(5) highlights that co-training the debiased classifier and semantic distribution detector further boosts performance. 
Notably, guiding the debiased learning with semantic distribution certainty and task consistency function yields a notable $4.6\%$ performance increase in Row~(6).

\noindent\textbf{Loss function.}
In addition, we explore the impact of the data and the respective loss functions employed during the training of debiased classifier, denoted as $\mathcal{L}_{adl}^s$ and $\mathcal{L}_{adl}^u$, targeting the labelled and unlabelled datasets, respectively. The results are shown in Tab.~\ref{tab:data}. 
These experiments are undertaken on the FGVC-Aircraft~\cite{maji2013fine} using various subset combinations. 
Solely training with $\mathcal{L}_{adl}^s$ introduces bias towards known categories, leading to a notable performance decline. 
Conversely, exclusive training with $\mathcal{L}_{adl}^u$ fails to reach optimal performance levels, underscoring the essential role of knowledge derived from labelled data. 
These outcomes demonstrate the vital significance of both $\mathcal{L}_{adl}^s$ and $\mathcal{L}_{adl}^u$ in optimizing the debiased classifier.

\begin{wraptable}{R}{0.5\textwidth}
\centering
\caption{Experimental results regarding threshold $\tau$ on the unlabelled set and validation set of FGVC-Aircraft~\cite{maji2013fine} dataset.}
\setlength{\tabcolsep}{3mm}{
\resizebox{0.5\textwidth}{!}{
\begin{tabular}{ccccccc}
        \toprule
        &\multicolumn{3}{c}{Unlabelled Set}&\multicolumn{3}{c}{Validation Set}\\
        \cmidrule(lr{1em}){2-4} \cmidrule(lr{1em}){5-7}
        $\tau$&All&Old&New&All&Old&New\\
        \midrule
        0.90&59.4&\textbf{64.7}&56.7 &58.9&61.1&56.8\\
        \rowcolor{my_blue} 0.85&\textbf{61.7}&63.9&\textbf{60.6} &\textbf{61.1}&\textbf{62.0}&\textbf{60.3}\\
        0.80&60.7&61.5&60.3 &60.6&61.6&59.6\\
        \bottomrule
        \end{tabular}
}
}
\label{tab:thr}
\end{wraptable}

\noindent\textbf{Debiasing threshold $\tau$.} 
Similar to self-training approaches~\cite{sohn2020fixmatch,zhang2021flexmatch}, the selection of the threshold for generating pseudo-labels also plays a crucial role in our approach. 
Consistent with the methods outlined in~\cite{wen2023parametric} and~\cite{vaze2022generalized}, we calibrate the threshold based on its performance on a separate validation set of the labelled data. Detailed results regarding different thresholds on the FGVC-Aircraft~\cite{wah2011caltech} dataset, covering performance on both the unlabelled training dataset and the validation set, are presented. As shown in Tab.~\ref{tab:thr}, the threshold is incrementally adjusted in intervals of $0.05$. Notably, the performance trends for both datasets align, with optimal performance achieved when the threshold is set to $0.85$.

\section{Conclusion}
\label{sec:con}
This paper presents DebGCD, a distribution-guided debiased learning framework for GCD, comprising three primary components. 
Firstly, we introduce an auxiliary debiased learning mechanism by concurrently training a parallel classifier with the GCD classifier, thereby facilitating optimization in the GCD feature space. Secondly, a semantic distribution detector is introduced to explicitly identify semantic shifts and implicitly enhance performance. Lastly, we propose a semantic distribution certainty score that enables a curriculum-based learning approach, promoting effective learning for both seen and unseen classes. Despite its simplicity, DebGCD showcases superior performance, as evidenced by comprehensive evaluation on seven public benchmarks.

\paragraph{Acknowledgements.}
This work is supported by National Natural Science Foundation of China (Grant No. 62306251),  Hong Kong Research Grant Council - Early Career Scheme (Grant No. 27208022), and HKU Seed Fund for Basic Research.

\bibliography{iclr2025_conference}
\bibliographystyle{iclr2025_conference}
\clearpage

\appendix
\section*{Appendix}
{\hypersetup{linkcolor=rose}
\startcontents[sections]
\printcontents[sections]{l}{1}{\setcounter{tocdepth}{2}}
}

\clearpage

\section{Additional Implementation Details}
We adopt the class splits of labelled (`Old') and unlabelled (`New') categories in~\cite{vaze2022generalized} for generic object recognition datasets (including CIFAR-10~\cite{krizhevsky2009learning} and CIFAR-100~\cite{krizhevsky2009learning}) and the fine-grained Semantic Shift Benchmark~\cite{vaze2022semantic} (comprising CUB~\cite{wah2011caltech}, Stanford Cars~\cite{krause20133d}, and FGVC-Aircraft~\cite{maji2013fine}). Specifically, for all these datasets except CIFAR-100, $50\%$ of all classes are selected as `Old' classes ($\mathcal{Y}_l$), while the remaining classes are treated as `New' classes ($\mathcal{Y}_u \backslash \mathcal{Y}_l$). For CIFAR-100, $80\%$ of the classes are designated as `Old' classes, while the remaining $20\%$ as `New' classes. 
Furthermore, for ImageNet-1K~\cite{deng2009imagenet}, which is not covered in~\cite{vaze2022generalized}, we follow~\cite{wen2023parametric} to select the first $500$ classes sorted by class ID as the labelled classes. 
For all the datasets, $50\%$ of the images from the labelled classes are randomly sampled to form the labelled dataset $\mathcal{D}_l$, and all remaining images are regarded as the unlabelled dataset $\mathcal{D}_u$. 
Moreover, following~\cite{vaze2022generalized} and~\cite{wen2023parametric}, the model's hyperparameters are chosen based on its performance on a hold-out validation set, formed by the original test splits of labelled classes in each dataset. 
All experiments utilize the PyTorch framework on a workstation with an Intel i7 CPU and eight Nvidia Tesla V100 GPUs. The models are trained with a batch size of 128 on a single GPU, except for the the model on CIFAR-100, ImageNet-100 and ImageNet-1K dataset, for which the training is performed with eight GPUs.

\clearpage

\section{Representation Visualization}
\begin{figure}[h]
\centering
\includegraphics[width = 0.7\textwidth]{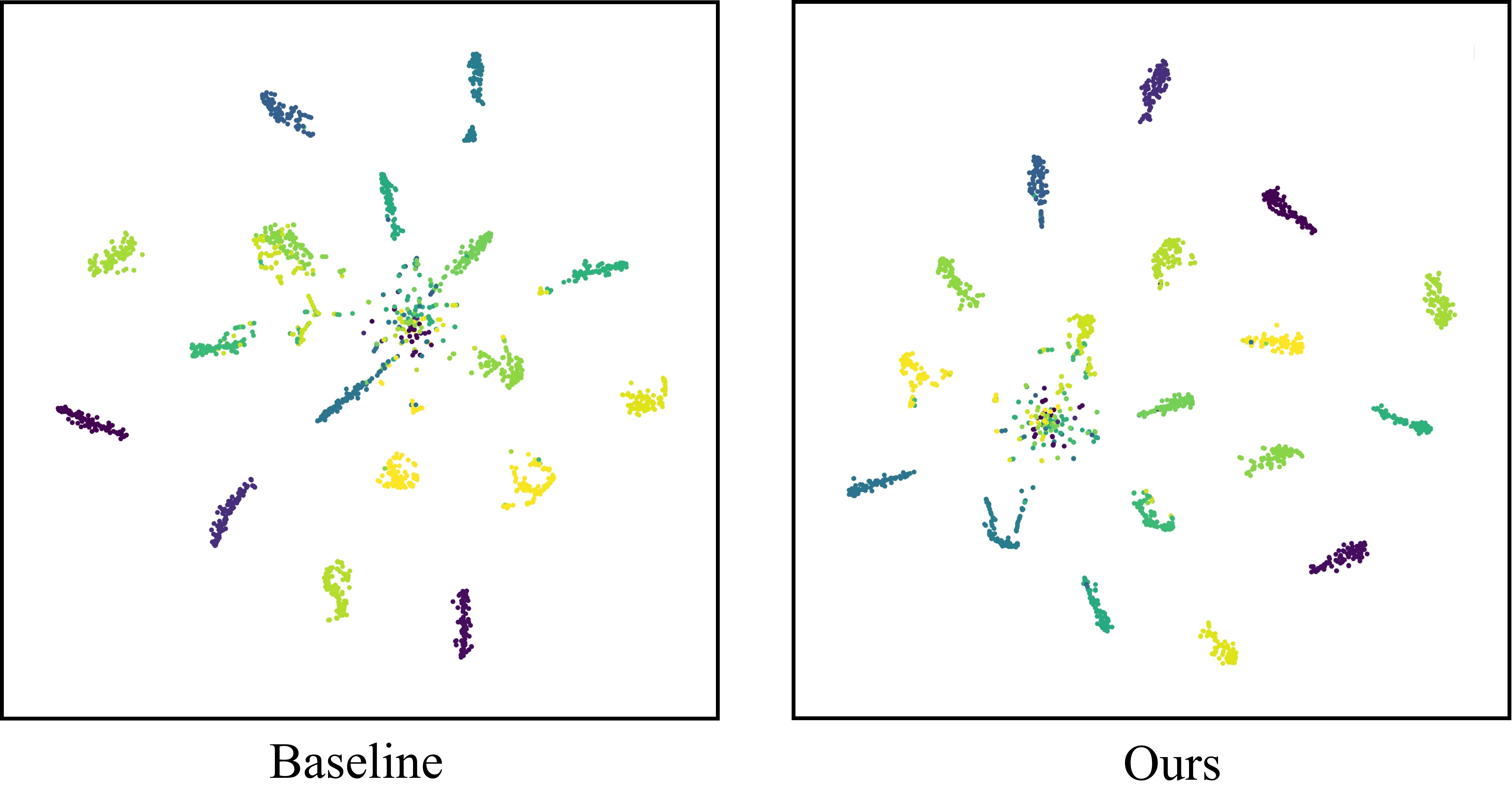}
\caption{
$t$-SNE visualization of 20 classes randomly sampled from the CIFAR-100~\cite{krizhevsky2009learning} dataset.
}
\label{fig:tsne}
\end{figure}

Here, we show the visual representation of the baseline and our method using $t$-SNE~\cite{van2008visualizing}. Specifically, we randomly select a set of $20$ classes, including $10$ from the `Old' categories and $10$ from the `New' categories. The clearly distinguishable clusters depicted in Fig.~\ref{fig:tsne} indicate that the features obtained within our framework form notably cohesive groupings compared to those of the baseline. This effectively demonstrates the optimization impacts induced by our method on the clustering feature space.

\clearpage

\section{Results on Additional Datasets}
To assess the performance of the proposed method comprehensively, we conducted evaluations on two more fine-grained datasets: Oxford-Pet~\cite{parkhi2012cats} and Herbarium 19~\cite{tan2019herbarium}. Oxford-Pet is a challenging dataset featuring various species of cats and dogs with limited data. 
Herbarium19, on the other hand, is a botanical research dataset encompassing diverse plant types, known for its long-tailed distribution and fine-grained categorization. 
The outcomes of our experiments on these datasets are detailed in Tab.~\ref{tab:add}. 
The results of SimGCD~\cite{wen2023parametric}  on Oxford-Pet are obtained through the execution of the officially released code. 
Our DebGCD model consistently demonstrates superior performance on both datasets. 

\begin{table*}[h]
\centering
\caption{Comparison with state-of-the-art GCD methods on Herbarium19~\cite{tan2019herbarium} and Oxford-Pet~\cite{parkhi2012cats}.} 
\setlength{\tabcolsep}{3mm}{
\resizebox{0.7\textwidth}{!}{
\begin{tabular}{l>{\columncolor{my_blue}}ccc>{\columncolor{my_blue}}ccc}
    \toprule
     &\multicolumn{3}{c}{Oxford-Pet}&\multicolumn{3}{c}{Herbarium19}\\
    \cmidrule(lr{1em}){2-4} \cmidrule(lr{1em}){5-7} 
 Method&All&Old&New&All&Old&New\\ 
    \hline
    $k$-means~\cite{macqueen1967some}&77.1&70.1&80.7 &13.0&12.2&13.4 \\
    RankStats+~\cite{han2021autonovel}&-&-&- &27.9&55.8&12.8 \\
    UNO+~\cite{fini2021unified}&-&-&- &28.3&53.7&14.7 \\
    ORCA~\cite{cao2022open}&-&-&- &24.6&26.5&23.7 \\
    GCD~\cite{vaze2022generalized}&80.2&85.1&77.6 &35.4&51.0&27.0\\
    XCon~\cite{fei2022xcon}&86.7&\underline{91.5}&84.1 &-&-&-\\
    OpenCon~\cite{sun2022opencon}&-&-&- &39.3&58.9&28.6\\
    DCCL~\cite{pu2023dynamic}&88.1&88.2&88.0 &-&-&- \\
    SimGCD~\cite{wen2023parametric}&\underline{91.7}&83.6&\underline{96.0} &44.0&58.0&36.4 \\
    $\mu$GCD~\cite{vaze2023no}&-&-&- &\textbf{45.8}&\textbf{61.9}&\textbf{37.2} \\
    InfoSieve~\cite{rastegar2023learn}&90.7&\textbf{95.2}&88.4 &40.3&59.0&30.2 \\ 
    \hline
\textbf{DebGCD }&\textbf{93.0}&86.4&\textbf{96.5} &\underline{44.7}&\underline{59.}4&\underline{36.8}\\
    \bottomrule
\end{tabular}
}
}
\label{tab:add}
\end{table*}

\clearpage

\section{Experiments with the Stronger DINOv2 Representations}
To further evaluate the robustness of the proposed method, we also evaluate the performance of DebGCD utilizing the stronger DINOv2~\cite{oquab2023dinov2} pre-trained weights. 
Like in~\cite{vaze2023no}, in Tab.~\ref{tab:dinov2}, we also compare our method with the k-means~\cite{macqueen1967some} baseline, and SimGCD~\cite{wen2023parametric}, $\mu$GCD~\cite{vaze2023no}. 
Our method outperforms other methods on CUB~\cite{wah2011caltech} and FGVC-Aircraft~\cite{maji2013fine} on `All', `Old' and `New' classes consistently. 
On Stanford Cars~\cite{krause20133d}, our method outperforms other methods on `New' classes,  while performing the second-best on `All' and `Old' classes. 
Moreover, for the average performance of `All' classes across the three datasets, DebGCD outperforms the SimGCD baseline by about $6\%$ and $\mu$GCD by about $3\%$. 
Additionally, we also evaluate our model on generic datasets and compare it with the SimGCD baseline in Tab.~\ref{tab:generic_v2}, demonstrating consistent improvement. 
The results on both fine-grained and generic datasets validate the robustness of our proposed method on the stronger DINOv2 representations, further showcasing its effectiveness.

\begin{table*}[h]
\centering
\caption{Comparison with state-of-the-art GCD methods on SSB leveraging DINOv2~\cite{oquab2023dinov2} pre-trained weights.}
\setlength{\tabcolsep}{2mm}{
\resizebox{0.9\textwidth}{!}{
\begin{tabular}{l>{\columncolor{my_blue}}ccc>{\columncolor{my_blue}}ccc>{\columncolor{my_blue}}ccc>{\columncolor{my_blue}}c>{\columncolor{my_blue}}c}
    \toprule
     &\multicolumn{3}{c}{CUB}&\multicolumn{3}{c}{Stanford Cars}&\multicolumn{3}{c}{FGVC-Aircraft}&\multicolumn{1}{c}{Average}\\
    \cmidrule(lr{1em}){2-4} \cmidrule(lr{1em}){5-7} \cmidrule(lr{1em}){8-10}
 Method&All&Old&New&All&Old&New&All&Old&New&All\\ \hline
    $k$-means~\cite{macqueen1967some}&67.6&60.6&71.1 &29.4&24.5&31.8 &18.9&16.9&19.9 &38.6\\
    GCD~\cite{vaze2022generalized}&71.9&71.2&72.3 &65.7&67.8&64.7 &55.4&47.9&59.2 &64.3\\
    CiPR~\cite{hao2023cipr}&\textbf{78.3}&73.4&\textbf{80.8} &66.7&77.0&61.8 &59.2&65.0&56.3 &68.1\\
    SimGCD~\cite{wen2023parametric}&71.5&\underline{78.1}&68.3 &71.5&81.9&66.6 &63.9&\underline{69.9}&60.9 &69.0\\
    $\mu$GCD~\cite{vaze2023no}&74.0&75.9&73.1 &\textbf{76.1}&\textbf{91.0}&\underline{68.9} &\underline{66.3}&68.7&\underline{65.1} &\underline{72.1}\\
    SPTNet~\cite{wang2024sptnet} &76.3&79.5&74.6 &-&-&- &-&-&- &-\\
    \hline
    \textbf{DebGCD }&\underline{77.5}&\textbf{80.8}&\underline{75.8} &\underline{75.4}&\underline{87.7}&\textbf{69.5} &\textbf{71.9}&\textbf{76.0}&\textbf{69.8} &\textbf{74.9}\\
    \bottomrule
\end{tabular}
}
}
\label{tab:dinov2}
\end{table*}

\begin{table*}[h]
\centering
\caption{
Comparison with state-of-the-art GCD methods on generic datasets leveraging DINOv2~\cite{oquab2023dinov2} pre-trained weights.}
\setlength{\tabcolsep}{2mm}{
\resizebox{0.95\textwidth}{!}{
\begin{tabular}{l>{\columncolor{my_blue}}ccc>{\columncolor{my_blue}}ccc>{\columncolor{my_blue}}ccc>{\columncolor{my_blue}}ccc>{\columncolor{my_blue}}ccc}
    \toprule
     &\multicolumn{3}{c}{CIFAR-10}&\multicolumn{3}{c}{CIFAR-100}&\multicolumn{3}{c}{ImageNet-100}&\multicolumn{3}{c}{ImageNet-1K}\\
    \cmidrule(lr{1em}){2-4} \cmidrule(lr{1em}){5-7} \cmidrule(lr{1em}){8-10} \cmidrule(lr{1em}){11-13}
 Method&All&Old&New&All&Old&New&All&Old&New &All&Old&New \\ \hline
    GCD~\cite{vaze2022generalized}&97.8&\textbf{99.0}&97.1 &79.6&84.5&69.9 &78.5&89.5&73.0 &-&-&- \\
    CiPR~\cite{hao2023cipr}&\textbf{99.0}&\underline{98.7}&99.2 &\textbf{90.3}&89.0&\textbf{93.1} &88.2&87.6&\underline{88.5} &-&-&-    \\
    SimGCD~\cite{wen2023parametric}  &98.7&96.7&\textbf{99.7} &88.5&\underline{89.2}&87.2 &89.9&95.5&87.1 &\underline{58.0}&\underline{66.9}&\underline{53.2}\\
    SPTNet~\cite{wang2024sptnet} &-&-&- &-&-&- &\underline{90.1}&\underline{96.1}&87.1 &-&-&-\\
    \textbf{DebGCD } &\underline{98.9}&97.5&\underline{99.6} &\underline{90.1}&\textbf{90.9}&\underline{88.6} &\textbf{93.2}&\textbf{97.0}&\textbf{91.2} &\textbf{71.7}&\textbf{86.2}&\textbf{64.5}\\
    \bottomrule
\end{tabular}
}
}
\label{tab:generic_v2}
\end{table*}

\clearpage

\section{Category Discovery with Estimated Category Numbers}
Following the majority of the literature, we experiment mainly using the ground-truth category numbers. 
In this section, we report the results of DebGCD using the number of categories estimated utilizing an off-the-shelf method~\cite{vaze2022generalized}, to showcase the performance with the ground-truth category numbers are not available. Tab.~\ref{tab:estk} reports the estimated numbers. We compare DebGCD with SimGCD~\cite{wen2023parametric}, $\mu$GCD~\cite{vaze2023no}, and GCD~\cite{vaze2022generalized} in Tab.~\ref{tab:estk1}. 
For both CUB~\cite{wah2011caltech} and Stanford Cars~\cite{krause20133d}, despite a discrepancy of approximately $15\%$ between the ground-truth and estimated category numbers, our method exhibits a smaller decline in performance compared to GCD and SimGCD. 
The same trend is also observed on Imagenet-100~\cite{deng2009imagenet}. 
DebGCD remains the most competitive method on `All' classes using the same estimated category numbers on all four datasets, which clearly demonstrates the robustness and effectiveness of our proposed method.

\begin{table}[h]
\centering
\caption{Estimated class numbers in the unlabelled data using method proposed in~\cite{vaze2022generalized}.}
\setlength{\tabcolsep}{1.5mm}{
\resizebox{0.55\textwidth}{!}{
\begin{tabular}{lccccc}
    \toprule
    &CUB&Stanford Cars&CIFAR-100&ImageNet-100\\
    \midrule
    Ground-truth $K$&200&196&100&100\\
    Estimated $K$&231&230&100&109\\
    \bottomrule
\end{tabular}
}
}
\label{tab:estk}
\end{table}

\begin{table}[h]
\centering
\caption{Results with the estimated number of categories. The estimated class numbers in Tab.~\ref{tab:estk} are adopted for all methods.
}

\setlength{\tabcolsep}{2mm}{
\resizebox{0.9\textwidth}{!}{
\begin{tabular}{l>{\columncolor{my_blue}}ccc>{\columncolor{my_blue}}ccc>{\columncolor{my_blue}}ccc>{\columncolor{my_blue}}ccc}
    \toprule
     &\multicolumn{3}{c}{CUB}&\multicolumn{3}{c}{Stanford Cars}&\multicolumn{3}{c}{CIFAR-100}&\multicolumn{3}{c}{ImageNet-100}\\
    \cmidrule(lr{1em}){2-4} \cmidrule(lr{1em}){5-7} \cmidrule(lr{1em}){8-10} \cmidrule(lr{1em}){11-13}
 Method&All&Old&New&All&Old&New&All&Old&New&All&Old&New\\ \hline
    GCD~\cite{vaze2022generalized}&47.1&55.1&44.8 &35.0&56.0&24.8 &73.0&76.2&66.5 &72.7&91.8&63.8\\
    SimGCD~\cite{wen2023parametric}&61.5&66.4&59.1 &49.1&65.1&41.3 &80.1&81.2&77.8 &81.7&91.2&76.8\\
    $\mu$GCD~\cite{vaze2023no}&62.0&60.3&\textbf{62.8} &56.3&66.8&51.1 &-&-&- &-&-&-\\
    \hline
    \textbf{DebGCD}&\textbf{64.5}&\textbf{68.5}&62.5 &\textbf{63.3}&\textbf{78.6}&\textbf{55.8} &\textbf{83.0}&\textbf{84.6}&\textbf{79.9} &\textbf{84.9}&\textbf{93.3}&\textbf{80.7}\\
    \bottomrule
\end{tabular}
}
}
\label{tab:estk1}
\end{table}

\clearpage

\section{Utilization Ratio of Unlabelled Data}
\begin{figure*}[h]
    \centering
    \subfigure{\label{fig:a}\includegraphics[width=67mm]{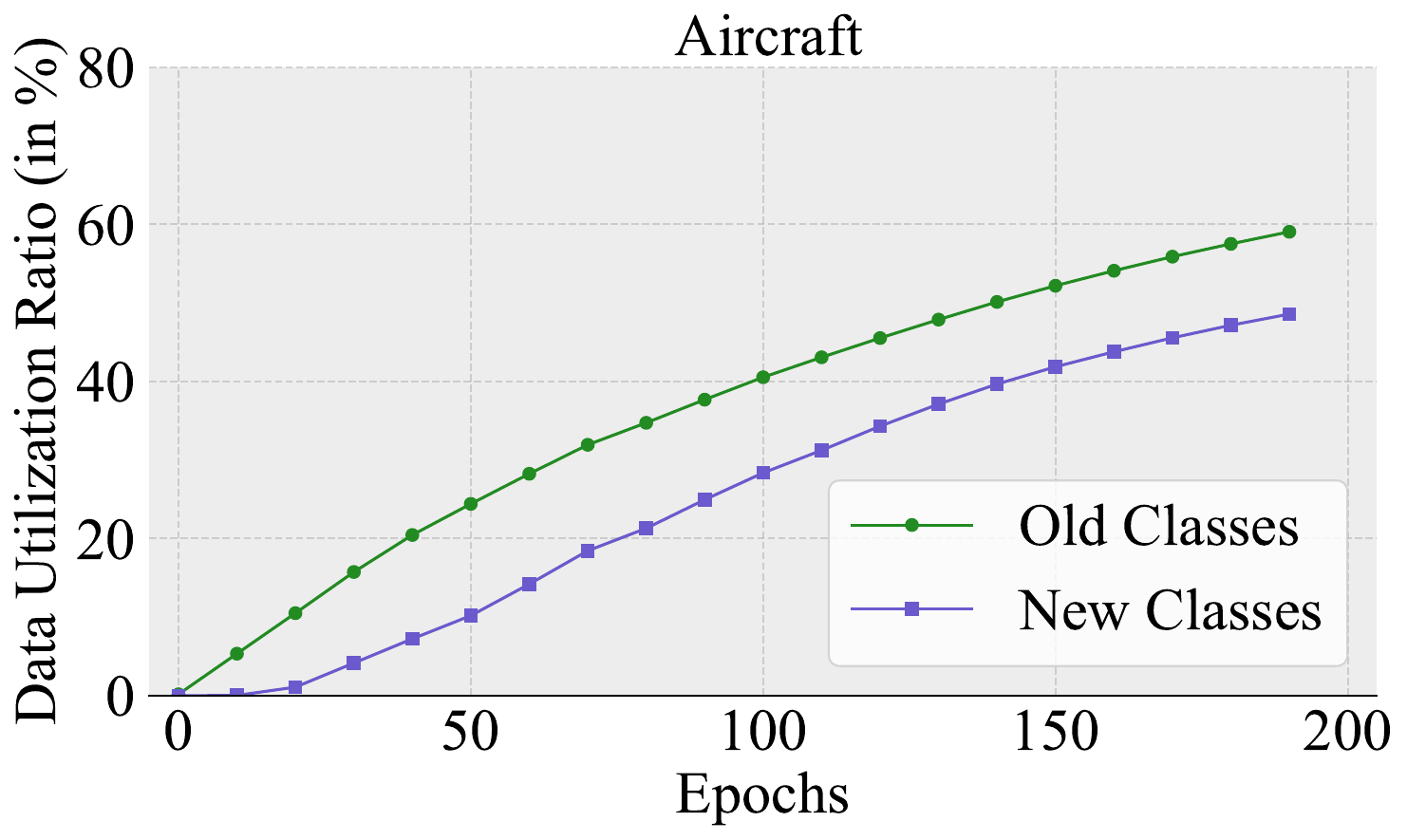}}
    \subfigure{\label{fig:b}\includegraphics[width=67mm]{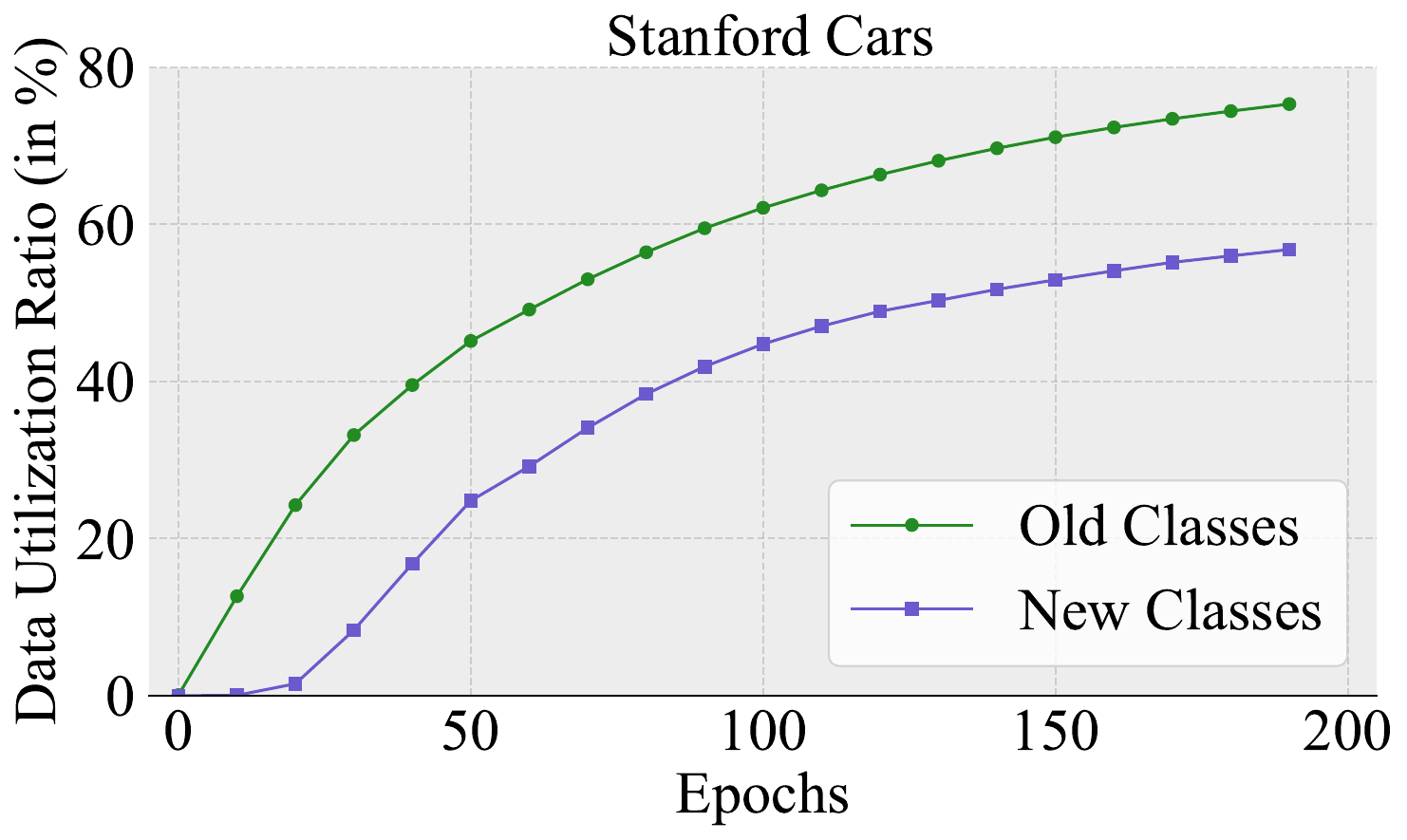}}
    \caption{Unlabelled data utilization ratios for `Old' and `New' classes during training on FGVC-Aircraft~\cite{maji2013fine} (left) and Stanford Cars~\cite{krause20133d} (right) datasets.}
    \label{fig:data}
\end{figure*}

The data utilization ratio is a notable index for pseudo-labeling methods, offering clear insights into the data efficiency. Our examination encompasses the utilization ratio of unlabelled data from both the `Old' and `New' classes during the training of the debiased classifier on FGVC-Aircraft~\cite{maji2013fine} and Stanford Cars~\cite{krause20133d}, as depicted in Fig.~\ref{fig:data}. Initially, the majority of data from the unknown categories remains untapped. Subsequently, after approximately 20 epochs, samples from unknown categories start to be incorporated. The utilization ratio keeps growing, reaching a ratio of around $40\%$ at the 100th epoch. Ultimately, more than $60\%$ of the known categories' samples and nearly half of the unknown categories' samples are utilized.

\clearpage

\section{GCD Classifier \textit{vs.} Debiased Classifer} 
\begin{figure}[h]
\centering
\includegraphics[width = \textwidth]{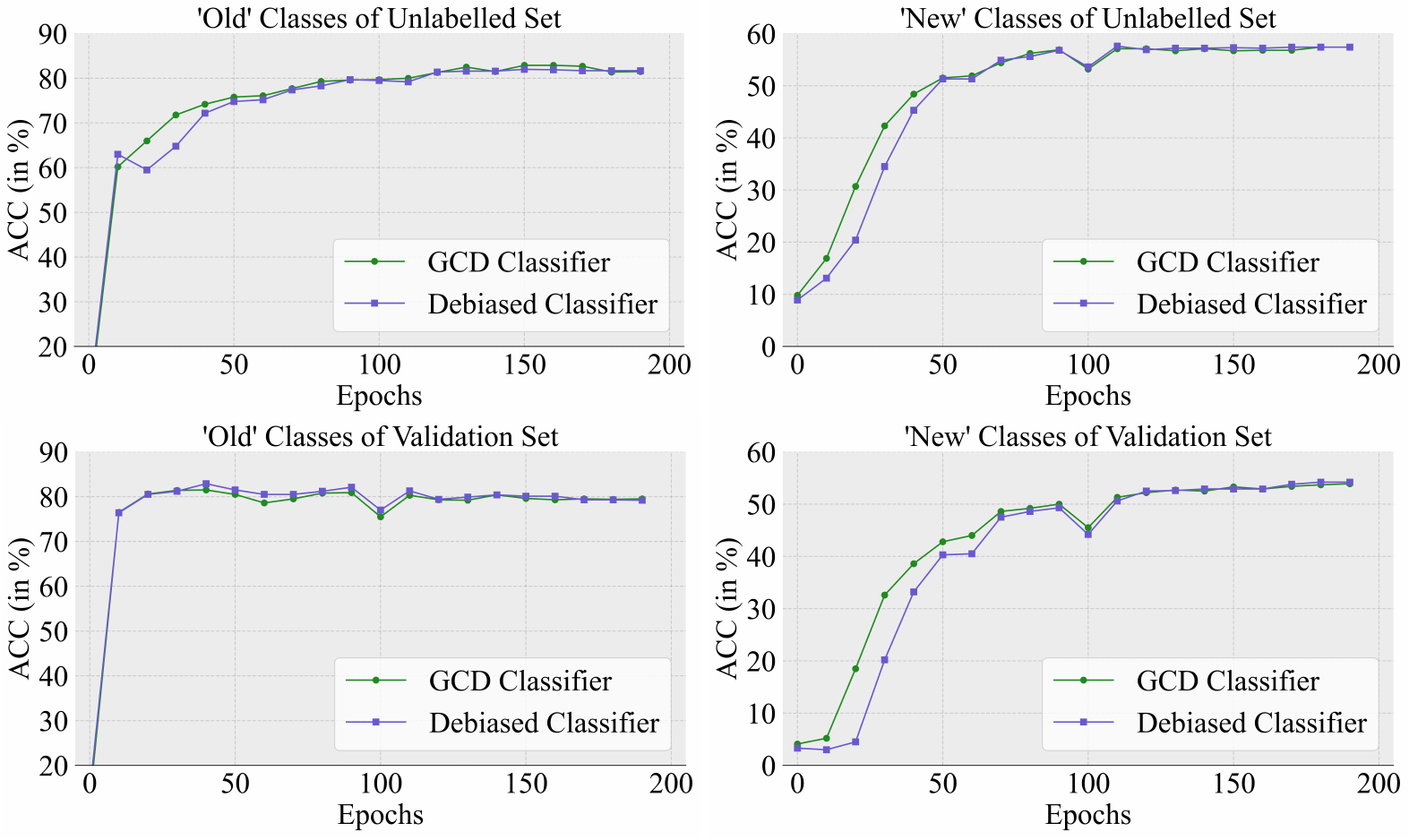}
\caption{
ACC evolution on both the `Old' and `New' classes of GCD Classifier and debiased classifier during training on Stanford Cars dataset~\cite{krause20133d}. 
The top two figures depict ACC on the unlabelled training set, while the bottom two illustrate ACC on the validation set.
}
\label{fig:classifier}
\end{figure}

We compare the performance between the two classifiers, the GCD Classifier and the debiased classifier, in our framework. We report the ACC results across different epochs in Fig.~\ref{fig:classifier} when training on Stanford Cars~\cite{krause20133d}, including unlabelled data from both training and the validation splits of the original dataset. Initially, the debiased classifier exhibits bias towards the `Old' classes, given that the training data primarily comprises labelled data from known categories. However, as predicted scores of the unlabelled samples, particularly those from the unknown categories, progressively surpass the debiasing threshold, the performance on the unknown categories gradually improves and eventually matches with the labelled categories. 
Ultimately, upon convergence of the model, the performance on both known and unknown categories converges to that of the GCD classifier.

\clearpage

\section{Performance of the Semantic Distribution Detector}
\begin{figure}[h]
\centering
\includegraphics[width = 0.95\textwidth]{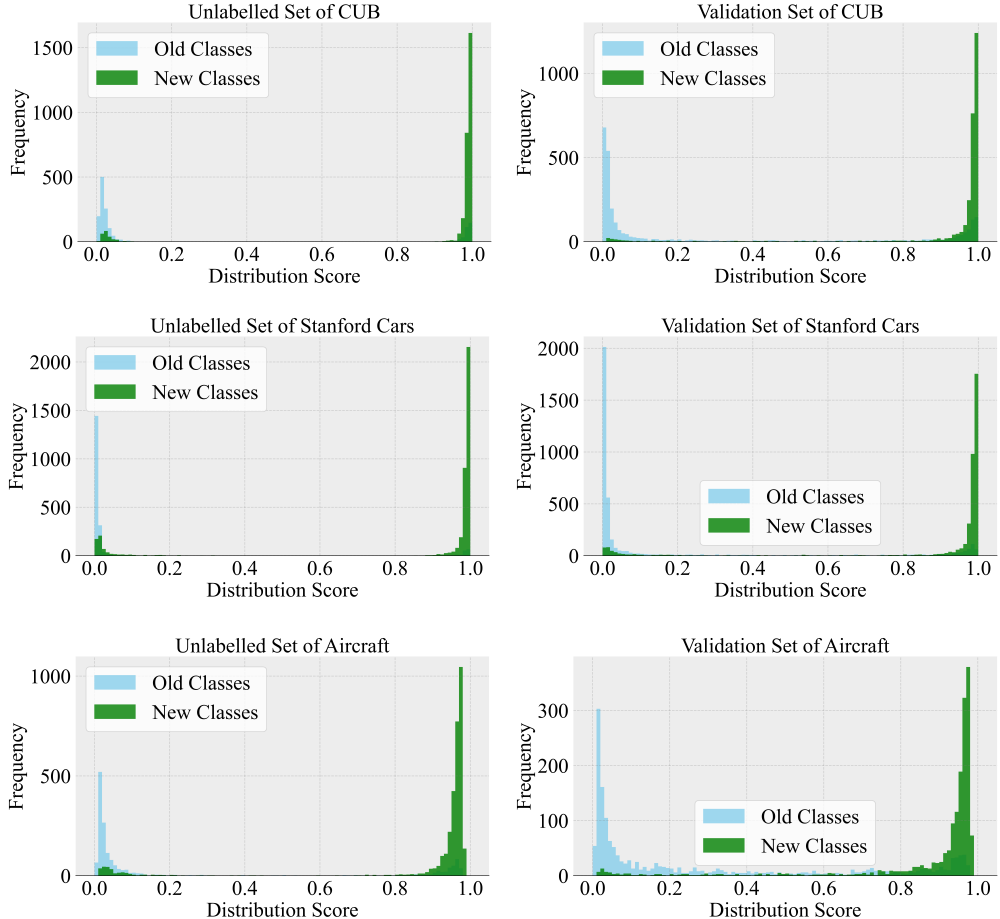}
\caption{
Histograms of the distribution scores $s_i$ for datasets in SSB~\cite{vaze2022semantic}.
}
\label{fig:ood}
\end{figure}

We evaluate the OOD detection performance of our semantic distribution detector in DebGCD, using the threshold-free Area Under the Receiver-Operator curve (AUROC) as the evaluation metric, which is widely used in the OOD detection literature. A comparison of the OOD performance between training the entire framework and training solely the distribution detector is presented in Tab.~\ref{tab:auroc}.  A significant improvement in OOD performance is obtained by training jointly the GCD classifier and debiased classifier. This aligns with the results presented in Tab.~4 of the main paper, which demonstrate the mutual benefits among the three branches (tasks) in our framework. 
Additionally, we visualize the distribution of the score $s_i$ on the challenging SSB datasets in Fig.~\ref{fig:ood} which shows that our method can successfully distinguish samples from `Old' and `New' classes in the unlabelled data of both the training and validation splits of the original dataset.

\begin{table}[h]
\centering
\caption{OOD performance in terms of AUROC on unlabelled data, including CIFAR-10~\cite{krizhevsky2009learning}, CIFAR-100~\cite{krizhevsky2009learning}, ImageNet-100~\cite{deng2009imagenet}, CUB~\cite{wah2011caltech}, Stanford Cars~\cite{krause20133d}, and FGVC-Aircraft~\cite{maji2013fine}.}
\resizebox{0.95\textwidth}{!}{
\begin{tabular}{lcccccc}
    \toprule
&CIFAR-10&CIFAR-100&ImageNet-100&CUB&Stanford Cars&FGVC-Aircraft\\
    \midrule
    $\mathcal{L}_{sdl}$&66.1&90.8&96.5&77.5&78.6&76.2\\
\rowcolor{my_blue}$\mathcal{L}_{sdl}$+$\mathcal{L}_{gcd}$+$\mathcal{L}_{adl}$&\textbf{97.5}&\textbf{94.8}&\textbf{99.5}&\textbf{86.8}&\textbf{89.6}&\textbf{86.3} \\
    \bottomrule
\end{tabular}
}
\label{tab:auroc}
\end{table}

\clearpage

\section{Analysis of Attention Maps}
\begin{figure}[ht]
\centering
\includegraphics[width = 0.9\textwidth]{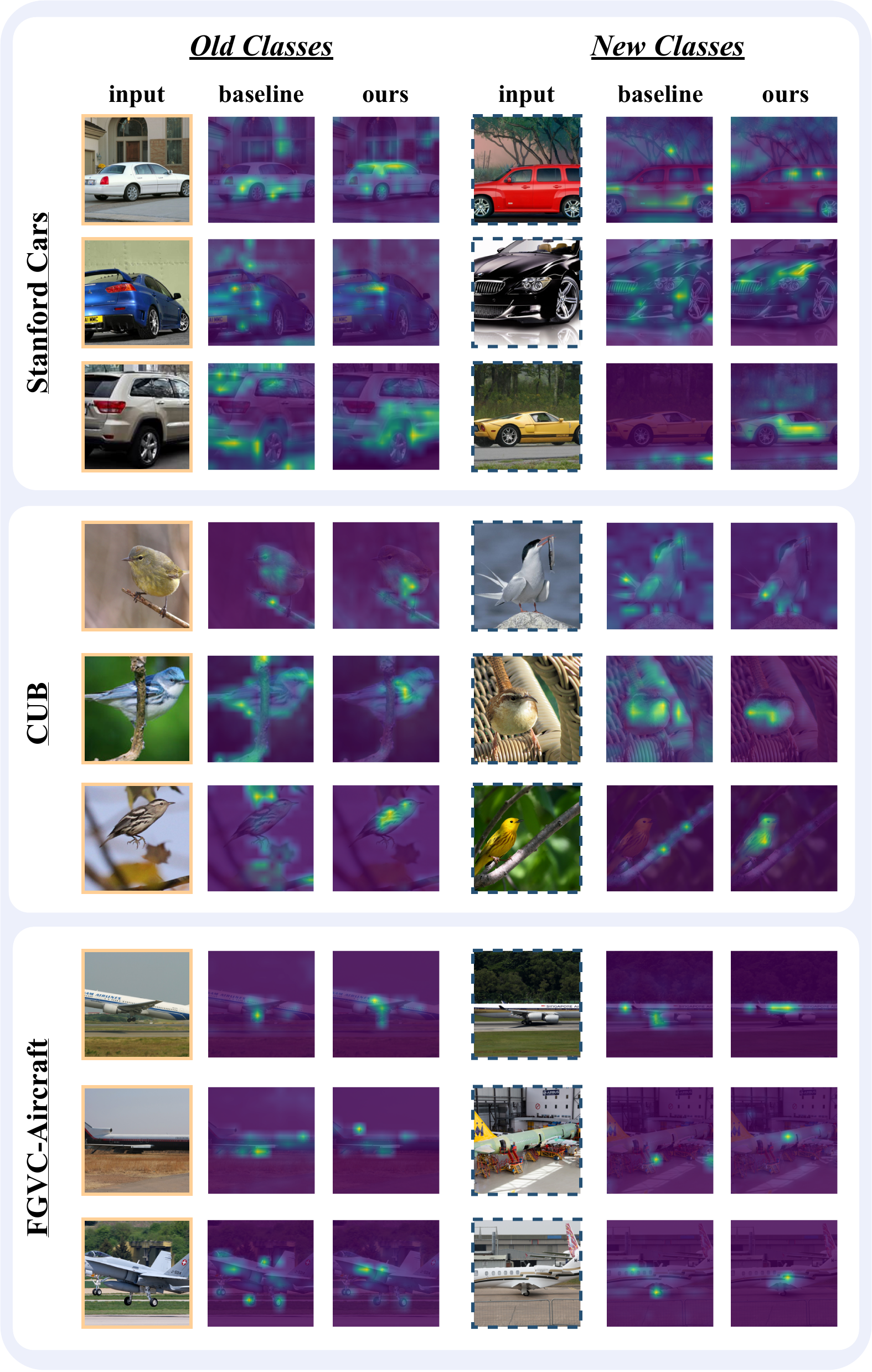}
\caption{
Visualization of attention maps. 
Our method successfully directs its attention towards foreground objects, irrespective of whether they belong to the `Old' or `New' classes. The baseline denotes the pre-trained DINO.
}
\label{fig:attn}
\end{figure}

In our DebGCD framework, both the backbone embedding space and the GCD classifier are optimized. Thus, the \texttt{CLS} token is indirectly optimized. We can glean insights from its attention with the patch embeddings. 
In Fig.~\ref{fig:attn}, we visualize the attention maps from the final transformer block in the DINO backbone~\cite{caron2021emerging} on the three fine-grained datasets in SSB benchmark~\cite{vaze2022semantic}. 
Within this final block, a multi-head self-attention layer with $12$ attention heads attends to the input features, producing $12$ attention maps between the \texttt{CLS} token and patch embeddings at a resolution of $14\times 14$. 
Following~\cite{caron2021emerging}, we compute the mean value of these attention maps and upsample them to the image size to visualize the most prominent regions. The visualization demonstrates that the attention maps generated by our model predominantly focus on the object of interest, effectively ignoring spurious factors and background clutter, while those of the DINO baseline are more scattered over the entire image.

\clearpage

\section{Ablation Studies on More Datasets}
In addition to the Stanford Cars dataset, we present ablation results on additional datasets to validate the effectiveness of the proposed components. These include the other two datasets from the SSB benchmark: CUB~\cite{wah2011caltech} and FGVC-Aircraft~\cite{maji2013fine}, as well as the generic dataset ImageNet-100~\cite{deng2009imagenet}, detailed in Tab.~\ref{tab:ablation2}. 
The results indicate that directly applying debiased learning to the original GCD classifier results in a performance decline across all three datasets (Row (1) \textit{vs.} Row (2)). In contrast, utilizing an auxiliary classifier leads to performance improvements of $3.3\%$, $3.5\%$, and $1.7\%$ on the three datasets, respectively, as observed in Row  (1) \textit{vs.} Row  (3). This underscores the importance of the auxiliary classifier in achieving effective debiased learning. 
Moreover, the joint training of the debiased classifier and the OOD detector provides further enhancements (Row (3) \textit{vs.} Row (5)). Lastly, the incorporation of distribution guidance results in additional performance improvements. These findings align with those observed on the Stanford Cars dataset, as demonstrated in manuscript.

\begin{table}[h]
\centering
\caption{Ablations on more datasets, including CUB~\cite{wah2011caltech}, FGVC-Aircraft~\cite{maji2013fine} and ImageNet-100~\cite{deng2009imagenet}. ACC of `All', `Old' and `New' categories are listed.}
\setlength{\tabcolsep}{2mm}{
\resizebox{1.0\textwidth}{!}{
\begin{tabular}{cccccccccccccc}
    \toprule
&\multirow{2}*{\shortstack{Debiased \\ Learning}}&\multirow{2}*{\shortstack{Auxiliary \\Classifier}}&\multirow{2}*{\shortstack{Semantic Dist. \\Learning}}&\multirow{2}*{\shortstack{Dist. \\Guidance}}&\multicolumn{3}{c}{CUB}&\multicolumn{3}{c}{FGVC-Aircraft}&\multicolumn{3}{c}{ImageNet-100}\\
\cmidrule(lr{1em}){6-8} \cmidrule(lr{1em}){9-11} \cmidrule(lr{1em}){12-14}
&&&& &All&Old&New &All&Old&New &All&Old&New\\
    \midrule
    (1)&\ding{55}&\ding{55}&\ding{55}&\ding{55} &60.3&65.6&57.7 &54.2&59.1&51.8 &83.0&93.1&77.9\\
    (2)&\ding{51}&\ding{55}&\ding{55}&\ding{55} &58.6&\bf{72.3}&51.7 &53.7&62.9&49.1 &82.8&94.1&77.2\\
    (3)&\ding{51}&\ding{51}&\ding{55}&\ding{55} &63.8&69.3&61.1 &57.7&59.8&56.5 &84.7&94.0&80.0\\
    (4)&\ding{55}&\ding{55}&\ding{51}&\ding{55} &61.3&69.4&57.3 &56.6&\bf{64.8}&52.5 &83.5&92.4&78.9\\
    (5)&\ding{51}&\ding{51}&\ding{51}&\ding{55} &64.9&70.9&61.9 &59.4&64.4&56.9 &85.0&93.8&80.3\\
    \rowcolor{my_blue} (6)&\ding{51}&\ding{51}&\ding{51}&\ding{51} &\bf{66.3}&71.8&\bf{63.5} &\bf{61.7}&63.9&\bf{60.6} &\bf{85.9}&\bf{94.3}&\bf{81.6} \\
    \bottomrule
\end{tabular}
}
}
\label{tab:ablation2}
\end{table}

\clearpage

\section{Impact of Hyperparameters}
In this section, we analyze the impact of hyperparameters in our DebGCD framework, including the depth of the projection network $\rho_s$, loss weights, and the number of tuned blocks.

\noindent\textbf{Depth of projection network $\rho_s$.}
As discussed in the paper, it is essential to disentangle the OOD and GCD feature spaces due to the differing learning objectives of these two tasks. To assess the impact of the depth of the projection network $\rho_s$, we conduct an experiment on the SSB benchmark, focusing on the number of layers in this MLP network. Here, a depth of $0$ denotes the absence of a projection network, meaning that the two tasks are optimized within the same feature space. 
As shown in Tab.~\ref{tab:mlp}, incorporating a $1$-layer $\rho_s$ results in performance improvements by $1.3\%$, $1.6\%$ and $1.1\%$ on CUB, Stanford Cars, and FGVC-Aircraft, respectively. The average GCD performance across all categories of DebGCD gradually improves as the number of MLP layers increases from $0$ to $5$. However, extending the MLP to $7$ layers yields little to no further improvement in performance. In our implementation, we therefore adopt a $5$-layer MLP for $\rho_s$ in our framework.

\noindent\textbf{Loss weights $\lambda_{sdl}$ and $\lambda_{adl}$.}
For these two loss weights, we first intuitively set the default value based on existing literature and our hypothesis. 
Our rationale for selecting values for the loss weights is as follows: 
For $\lambda_{sdl}$, we take inspiration from the previous literature using OVA classifier~\cite{saito2021ovanet}. In the paper, the model is fine-tuned with a learning rate of $10^{-3}$ , while the learning rate in the SimGCD baseline is $0.1$ (which is $100$ times larger than $10^{-3}$). To achieve a similar learning effect, as validated in~\cite{saito2021ovanet}, we scale our $\lambda_{sdl}$ value from $1.0$ down to $1/100$. Therefore, we set $\lambda_{sdl} = 0.01$ by default. 
For $\lambda_{adl}$, the weight of the debiased classifier, we expect it to play an important role similar to that of the original GCD classifier (where the loss weight is set to $1.0$). Thus, we have defaulted this value to $1.0$. 
After determining the default values, we conducted experiments on the SSB benchmark regarding the two loss weights by exploring values around the defaults. For $\lambda_{sdl}$, the range was ($0.005$, $0.01$, $0.02$). As for $\lambda_{adl}$, the range was ($0.5$, $1.0$, $2.0$). The impact of $\lambda_{sdl}$ is detailed below in Tab.~\ref{tab:lsdl}, with $\lambda_{adl}$ set to $1.0$. The impact of $\lambda_{adl}$ is illustrated below in Tab.~\ref{tab:ladl}, with $\lambda_{sdl}$ set to $0.01$. The results are in line with our hypothesis, indicating that our selected hyperparameters are indeed reasonable.

\begin{table}[h]
\centering
\caption{GCD performance on SSB~\cite{vaze2022semantic} using different number of layers in $\rho_s$.}
\setlength{\tabcolsep}{2mm}{
\resizebox{0.7\textwidth}{!}{
\begin{tabular}{cccccccccccccc}
    \toprule
     &\multicolumn{3}{c}{CUB}&\multicolumn{3}{c}{Stanford Cars}&\multicolumn{3}{c}{FGVC-Aircraft}&\multicolumn{1}{c}{Average}\\
    \cmidrule(lr{1em}){2-4} \cmidrule(lr{1em}){5-7} \cmidrule(lr{1em}){8-10} 
 MLP layer&All&Old&New&All&Old&New&All&Old&New&All\\ \hline
    0&63.6&\bf{75.2}&57.8 &62.3&76.2&54.1 &59.6&62.2&58.3 &61.8\\
    1&64.9&71.6&61.6 &63.9&80.2&56.0 &60.7&63.7&59.2 &63.1\\
    3&66.0&73.5&62.3 &64.7&\bf{82.2}&56.2 &61.1&64.2&59.5 &63.9\\
    \rowcolor{my_blue} 5&\bf{66.3}&71.8&\bf{63.5} &\bf{65.3}&81.6&\bf{57.4} &61.7&63.9&\bf{60.6} &\bf{64.4}\\
    7&65.8&72.0&62.7 &64.8&80.5&57.3 &\bf{61.9}&\bf{65.2}&60.3 &64.1\\
    \bottomrule
\end{tabular}
}
}
\label{tab:mlp}
\end{table}

\begin{table}[h]
\centering
\caption{GCD performance on SSB~\cite{vaze2022semantic} using different values of $\lambda_{sdl}$.}
\setlength{\tabcolsep}{2mm}{
\resizebox{0.65\textwidth}{!}{
\begin{tabular}{cccccccccccccc}
    \toprule
     &\multicolumn{3}{c}{CUB}&\multicolumn{3}{c}{Stanford Cars}&\multicolumn{3}{c}{FGVC-Aircraft}&\multicolumn{1}{c}{Average}\\
    \cmidrule(lr{1em}){2-4} \cmidrule(lr{1em}){5-7} \cmidrule(lr{1em}){8-10} 
 $\lambda_{sdl}$&All&Old&New&All&Old&New&All&Old&New&All\\ \hline
    0.02&65.5&\bf{73.2}&61.6 &64.3&79.2&57.1 &60.6&63.5&59.1 &63.5\\
    \rowcolor{my_blue} 0.01&\bf{66.3}&71.8&\bf{63.5} &\bf{65.3}&\bf{81.6}&\bf{57.4} &61.7&63.9&\bf{60.6} &\bf{64.4}\\
    0.005&65.8&72.4&62.5 &64.9&81.2&57.0 &\bf{62.1}&\bf{65.4}&60.3 &64.3\\
    \bottomrule
\end{tabular}
}
}
\label{tab:lsdl}
\end{table}

\begin{table}[h]
\centering
\caption{GCD performance on SSB~\cite{vaze2022semantic} using different values of $\lambda_{adl}$.}
\setlength{\tabcolsep}{2mm}{
\resizebox{0.65\textwidth}{!}{
\begin{tabular}{cccccccccccccc}
    \toprule
     &\multicolumn{3}{c}{CUB}&\multicolumn{3}{c}{Stanford Cars}&\multicolumn{3}{c}{FGVC-Aircraft}&\multicolumn{1}{c}{Average}\\
    \cmidrule(lr{1em}){2-4} \cmidrule(lr{1em}){5-7} \cmidrule(lr{1em}){8-10} 
 $\lambda_{adl}$&All&Old&New&All&Old&New&All&Old&New&All\\ \hline
    0.5&64.3&\bf{72.2}&60.3 &63.6&79.3&56.1 &60.2&63.5&58.6 &62.7 \\
    \rowcolor{my_blue} 1.0&\bf{66.3}&71.8&\bf{63.5} &\bf{65.3}&81.6&\bf{57.4} &\bf{61.7}&\bf{63.9}&\bf{60.6} &\bf{64.4} \\
    2.0&65.5&70.8&62.8 &64.1&\bf{83.0}&55.0 &60.4&63.5&58.8 &63.3 \\
    \bottomrule
\end{tabular}
}
}
\label{tab:ladl}
\end{table}

\clearpage
\noindent\textbf{Number of tuned blocks.}
In the baseline configuration~\cite{wen2023parametric}, only the last transformer block of the ViT-B/16 backbone is fine-tuned during training. In contrast, our framework incorporates additional tasks, including OOD detection and debiased learning, which would require different embedding spaces, thus calling for the need of more trainable parameters. 
In our experiments on both fine-grained and generic datasets, we explore tuning the last two blocks, and we note that tuning more than two blocks may lead to instability during training. Furthermore, we observe that increasing the number of tuned blocks can improve performance on specific datasets, particularly those that are fine-grained. 
As shown in Table~\ref{tab:tb}, tuning one additional transformer block leads to a performance improvement of over $1\%$ on the fine-grained datasets. In contrast, the performance enhancement on the generic datasets is more modest, at no more than $0.6\%$. 
Similar strategies have also been employed in previous methods, such as Infosieve~\cite{rastegar2023learn}.

\begin{table}[h]
\centering
\caption{GCD performance of SimGCD and DebGCD by tuning different numbers of transformer blocks.}
\setlength{\tabcolsep}{1.7mm}{
\resizebox{1.0\textwidth}{!}{
\begin{tabular}{lcccccccccccccccccccc}
    \toprule
     &&\multicolumn{3}{c}{CUB}&\multicolumn{3}{c}{Stanford Cars}&\multicolumn{3}{c}{FGVC-Aircraft}&\multicolumn{3}{c}{ImageNet-100}&\multicolumn{3}{c}{CIFAR-100}\\
    \cmidrule(lr{1em}){3-5} \cmidrule(lr{1em}){6-8} \cmidrule(lr{1em}){9-11} \cmidrule(lr{1em}){12-14} \cmidrule(lr{1em}){15-17}
 Method& \# of tuned blocks&All&Old&New&All&Old&New&All&Old&New&All&Old&New&All&Old&New\\ \hline
    SimGCD&1&60.3&65.6&57.7 &53.8&71.9&45.0 &54.2&59.1&51.8 &83.0&93.1&77.9 &80.1&81.2&77.8\\
    SimGCD&2& 60.8&65.8&58.4 & 53.6&67.6&49.8 & 52.8&56.8&50.8 & 83.2&92.9&78.3 & 79.4&80.1&77.3\\
    DebGCD&1&65.1&70.9&62.2 & 63.0&80.2&54.7 & 60.4&\textbf{65.0}&58.1 & 85.7&94.0&81.5 & 82.4&83.6&79.5\\
    \rowcolor{my_blue} DebGCD&2& \textbf{66.3}&\textbf{71.8}&\textbf{63.5} & \textbf{65.3}&\textbf{81.6}&\textbf{57.4} & \textbf{61.7}&63.9&\textbf{60.6} & \textbf{85.9}&\textbf{94.3}&\textbf{81.6} & \textbf{83.0}&\textbf{84.6}&\textbf{79.9}\\
    \bottomrule
\end{tabular}
}
}
\label{tab:tb}
\end{table}

\clearpage

\section{Stability Analysis}
Following the baseline established in~\cite{wen2023parametric}, we also assess the stability of the proposed method across all datasets utilized in our experiments. 
Tab.~\ref{tab:stability} reports the average results together with the standard deviations,  over three independent runs. 
Compared to the baseline results reported in~\cite{wen2023parametric}, we observe that the variance of DebGCD is even smaller, despite achieving significantly higher performance.

\begin{table}[h]
\centering
\caption{Complete results of DebGCD and SimGCD over three independent runs.}
\setlength{\tabcolsep}{2mm}{
\resizebox{0.7\textwidth}{!}{
\begin{tabular}{lccccccccc}
    \toprule
     &\multicolumn{3}{c}{SimGCD}&\multicolumn{3}{c}{DebGCD}\\
    \cmidrule(lr{1em}){2-4} \cmidrule(lr{1em}){5-7}  
 Dataset&All&Old&New&All&Old&New\\ 
 \hline
    CUB &60.3±0.1 &65.6±0.9 &57.7±0.4& 66.4±0.4 & 72.9±0.6 & 63.2±0.4 \\ 
    Stanford Cars &53.8±2.2 &71.9±1.7 &45.0±2.4 & 65.2±0.7 & 81.7±1.2 & 57.3±0.6 \\ 
    FGVC-Aircraft &54.2±1.9 &59.1±1.2 &51.8±2.3 & 61.7±0.5 & 65.9±1.2 & 59.5±1.1 \\ 
    CIFAR-10 &97.1±0.0 &95.1±0.1 &98.1±0.1& 97.3±0.1 & 95.0±0.2 & 98.4±0.1 \\ 
    CIFAR-100 &80.1±0.9 &81.2±0.4 &77.8±2.0 & 83.1±0.7 & 84.7±0.7 & 80.0±0.9 \\ 
    ImageNet-100 &83.0±1.2 &93.1±0.2 &77.9±1.9 & 86.1±0.6 & 94.5±0.5 & 81.8±0.6 \\ 
    ImageNet-1K  &57.1±0.1 &77.3±0.1 &46.9±0.2 & 64.9±0.3 & 82.1±0.2 & 56.4±0.4 \\ 
    Oxford-Pet &-&-&- & 93.2±0.2 & 86.3±0.1 & 96.8±0.3 \\ 
    Herbarium19 &44.0±0.4 &58.0±0.4 &36.4±0.8 & 44.9±0.3 & 59.3±0.3 & 37.1±0.5 \\ 
    \bottomrule
\end{tabular}
}
}
\label{tab:stability}
\end{table}

\clearpage

\section{Prediction Error Analysis}
In this section, we provide quantitative analysis on the improvements brought by our method from the perspective of prediction errors. 
Particularly, we examine the baseline model’s prediction by categorizing the errors into four types based on the relationship between the predicted (`Pred') and ground-truth (`GT') classes: `True Old', `False New', `False Old', and `True New'. 
`True Old' refers to incorrectly predicting an `Old' class sample as another `Old' class. `False New' indicates incorrectly predicting an `Old' class sample as a `New' class. Conversely, `False Old' means incorrectly predicting a `New' class sample as an `Old' class, and `True New' refers to incorrectly predicting a `New' class sample as another `New' class. 
From this perspective, our debiased learning method primarily aims to mitigate the label bias between `Old' and `New' classes, thereby reducing the likelihood of `New' class samples being predicted as `Old'. 
Consequently, this reduction in bias leads to a decrease in `False Old' predictions while reducing the errors of all the other three types.

\begin{figure}[h]
\centering
\includegraphics[width = 0.95\textwidth]{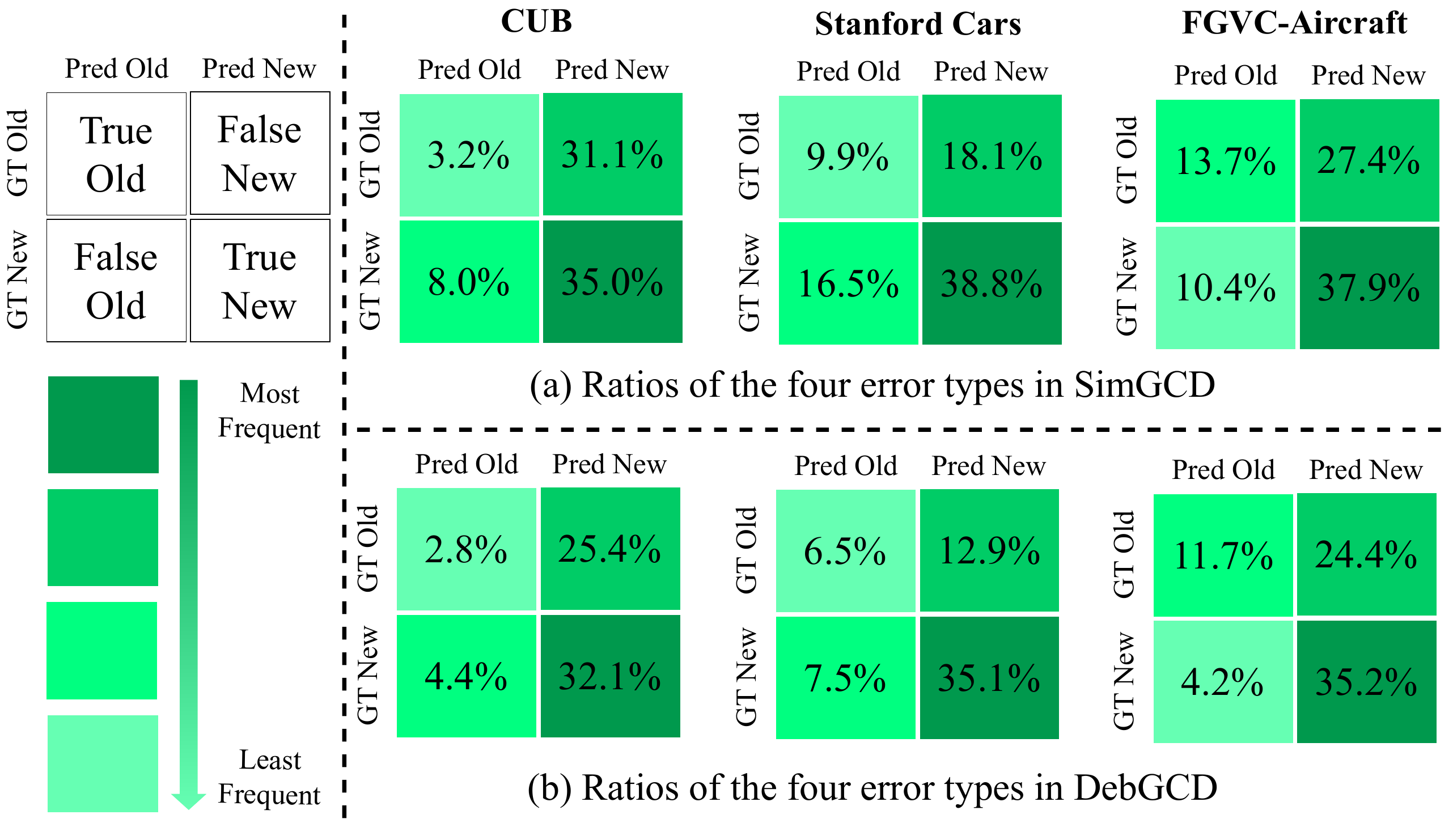}
\caption{
Ratios of the four types of prediction errors in GCD on SSB benchmark using SimGCD and DebGCD with DINO~\cite{caron2021emerging} pre-trained backbone. `Pred' and `GT' refer to the predicted and ground-truth results, respectively.
}
\label{fig:err_ana}
\end{figure}

In Fig.~\ref{fig:err_ana}, we present the ratios of the four types of \textit{prediction errors} as a proportion of the total number of samples in the new or old categories across three datasets in the SSB benchmark. 
As shown in Fig.~\ref{fig:err_ana}~(a), the error distributions vary significantly across datasets. 
Notably, the Stanford Cars dataset exhibits the highest number ($16.5\%$) of `False Old' samples, explaining why our method demonstrates the most substantial performance improvement on this dataset. 
In contrast, the CUB dataset shows the fewest ($8.0\%$) `False Old' samples, indicating relatively limited potential for performance enhancement. 
Comparing Fig.~\ref{fig:err_ana}~(a) and Fig.~\ref{fig:err_ana}~(b), we can see a significant reduction on the ratio of `False Old' as well as other three types of errors on all the three datasets.

\clearpage



\end{document}